\documentclass[review]{fcs}
\usepackage{multirow}
\usepackage{amsmath}
\usepackage{array}
\usepackage{setspace}
\usepackage{url}
\usepackage{graphicx}
\usepackage{epstopdf}
\usepackage{bm}
\usepackage{makecell}
\usepackage{ragged2e}
\usepackage{booktabs}

\title{The Future of Cognitive Strategy-enhanced Persuasive Dialogue Agents: New Perspectives and Trends}
\shorttitle{}
\author[1]{Mengqi Chen}
\author*[1]{Bin Guo}
\author[1]{Hao Wang}
\author[1]{Haoyu Li}
\author[1]{Qian Zhao}
\author[1]{Jingqi Liu}
\author[1]{Yasan Ding}
\author[2]{Yan Pan}
\author[1]{Zhiwen Yu}

\address[1]{Northwestern Polytechnical University, Xi'an, China}
\address[2]{National University of Defense Technology, Changsha, China}

\fcssetup{
  received       = {month dd, yyyy},
  accepted       = {month dd, yyyy},
  corr-email     = {guob@nwpu.edu.cn},
}
\begin{abstract}
Persuasion, as one of the crucial abilities in human communication, has garnered extensive attention from researchers within the field of intelligent dialogue systems. 
Developing dialogue agents that can persuade others to accept certain standpoints is essential to achieving truly intelligent and anthropomorphic dialogue systems.
Benefiting from the substantial progress of Large Language Models (LLMs), dialogue agents have acquired an exceptional capability in context understanding and response generation. 
However, as a typical and complicated cognitive psychological sy\-stem, persuasive dialogue agents also require knowledge from the domain of cognitive psychology to attain a level of human-like persuasion.
Consequently, the cognitive strategy-enhanced persuasive dialogue agent (defined as \textbf{\textit{CogAgent}}), which incorporates cognitive strategies to achieve persuasive targets through conversation, has become a predominant research paradigm. 
To depict the research trends of CogAgent, in this paper, we first present several fundamental cognitive psychology theories and give the formalized definition of three typical cognitive strategies, including the persuasion strategy, the topic path planning strategy, and the argument structure prediction strategy. 
Then we propose a new system architecture by incorporating the formalized definition to lay the foundation of CogAgent. 
Representative works are detailed and investigated according to the combined cognitive strategy, followed by the summary of authoritative benchmarks and evaluation metrics. Finally, we summarize our insights on open issues and future directions of CogAgent for upcoming researchers.
\end{abstract}

\begin{document}

\section{Introduction}
Dialog agents can engage in chitchat with humans to establish certain emotional connections or help us complete tasks through long-form conversations (e.g., restaurant reservation, travel time arrangement).
Building intelligent human-machine dialogue agen\-ts that can conduct natural and engaging conversations with humans is the long-standing goal of artificial intelligence (AI) \cite{guo2021conditional,huang2020challenges}. 
Moreover, the persuasive ability of dialogue agents has garnered extensive attention from researchers.
Persuasion is one of the crucial abilities in human communication. The Elaboration Likelihood Model (ELM) theory \cite{petty1986elaboration} suggests that people tend to engage with persuasive messages when communicating with others.
It is a prevalent phenomenon for individuals to hold diverse perspectives on a given topic and endeavor to influence others in altering their viewpoints, attitudes, or behaviors through conversational interactions \cite{fogg2002persuasive, ijsselsteijn2006persuasive}. 
There are massive persuasive scenarios in the real world, such as bargaining prices of goods, debating on specific topics, and arguing in online comment sections \cite{fogg2008mass, tan2016winning}, and active online communities, such as Debate\footnote{\url{https://www.debate.org/}} and ChangeMyView\footnote{\url{https://www.reddit.com/r/changemyview/}}, for people to communicate and influence other people's opinions by posting their views \cite{hidey2017analyzing}. 
A persuasive conversion includes two distinct parties, corresponding to persuader and persuadee, respectively \cite{torning2009persuasive}. The goal of the persuader is to change the persuadee's viewpoint on a specific topic by combining cognitive strategies, the personality of the persuadee, and other context features \cite{eagly1984cognitive, wang2019persuasion}.
The development of intelligent persuasive dialogue agents that can persuade users to accept certain standpoints is emerging as a promising research field \cite{shi2020effects, joshi2021dialograph}.

Modern dialogue agents have arrived at the era characterized by large language models (LLMs) \cite{min2021recent, zhao2023survey}.
Driven by an immense scale of parameters and an abundance of training data, dialogue agents (e.g., ChatGPT\footnote{\url{https://openai.com/blog/chatgpt/}}, LLaMA \cite{touvron2023llama}, Cla\-ude \cite{bai2022constitutional}, ChatGLM\footnote{\url{https://github.com/THUDM/ChatGLM3}}) have acquired an exceptional capability of context understanding and response generation \cite{zhou2023comprehensive, ray2023chatgpt}, reaching a satisfactory level of fluency, logic, emotional expression and personalization when conversing with humans \cite{li2024generating,wang2022informative}.
In addition to engaging in casual conversations with humans, existing dialogue agents, represented by ChatGPT, can assist humans in accomplishing intricate tasks, such as writing codes \cite{vaithilingam2022expectation, ni2023lever}, writing long and coherent academic papers \cite{yuan2022wordcraft, dergaa2023human} and aiding in office works (e.g., Microsoft Copilot\footnote{\url{https://adoption.microsoft.com/en-us/copilot/}}), thereby substantially augmenting productivity and the quality of life.

However, the persuasion process is an activity that involves human psychological cognition \cite{eagly1984cognitive, bless1990mood, petty2015emotion}. The design of persuasive dialogue agents needs to incorporate cognitive strategies to organize content, logic, and presentation of dialogue response reasonably from the cognitive psychology perspective.
There have been considerable works of persuasive dialogue systems, which ma\-inly enhance persuasiveness from three aspects, na\-mely, integrating persuasion strategies, planning topic pa\-ths, and extracting argument structures.
For example, Wang \textit{et al.} \cite{wang2019persuasion} provide an in-depth analysis of the impact of persuasion strategies on the persuasive power of dialogue systems in the context of persuasion for donation scenarios. 
To enable persuasion efficiently, Qin \textit{et al.} \cite{qin2020dynamic} drive a conversation towards a specified persuasive target with explicit topics/keywords planning over kno\-wledge graphs. 
Arguments are also an important source of effective persuasive content. Prakken \textit{et al.} \cite{prakken2020persuasive} develop a dialogue system wherein an argument graph serves as the persuasive knowledge for persuasive response generation.
Different from existing works that focus on the persuasiveness of dialogue systems, in this paper, we argue that persuasion is a cognitive psychology activity and that the persuasion strategy, the topic path planning strategy, and the argument structure prediction strategy can all be categorized as cognitive strategies.
We define cognitive strategy-enhanced persuasive dialogue agent as \textbf{\textit{CogAgent}}.

\begin{figure}[t]
\centering
\includegraphics[width=0.85\columnwidth]{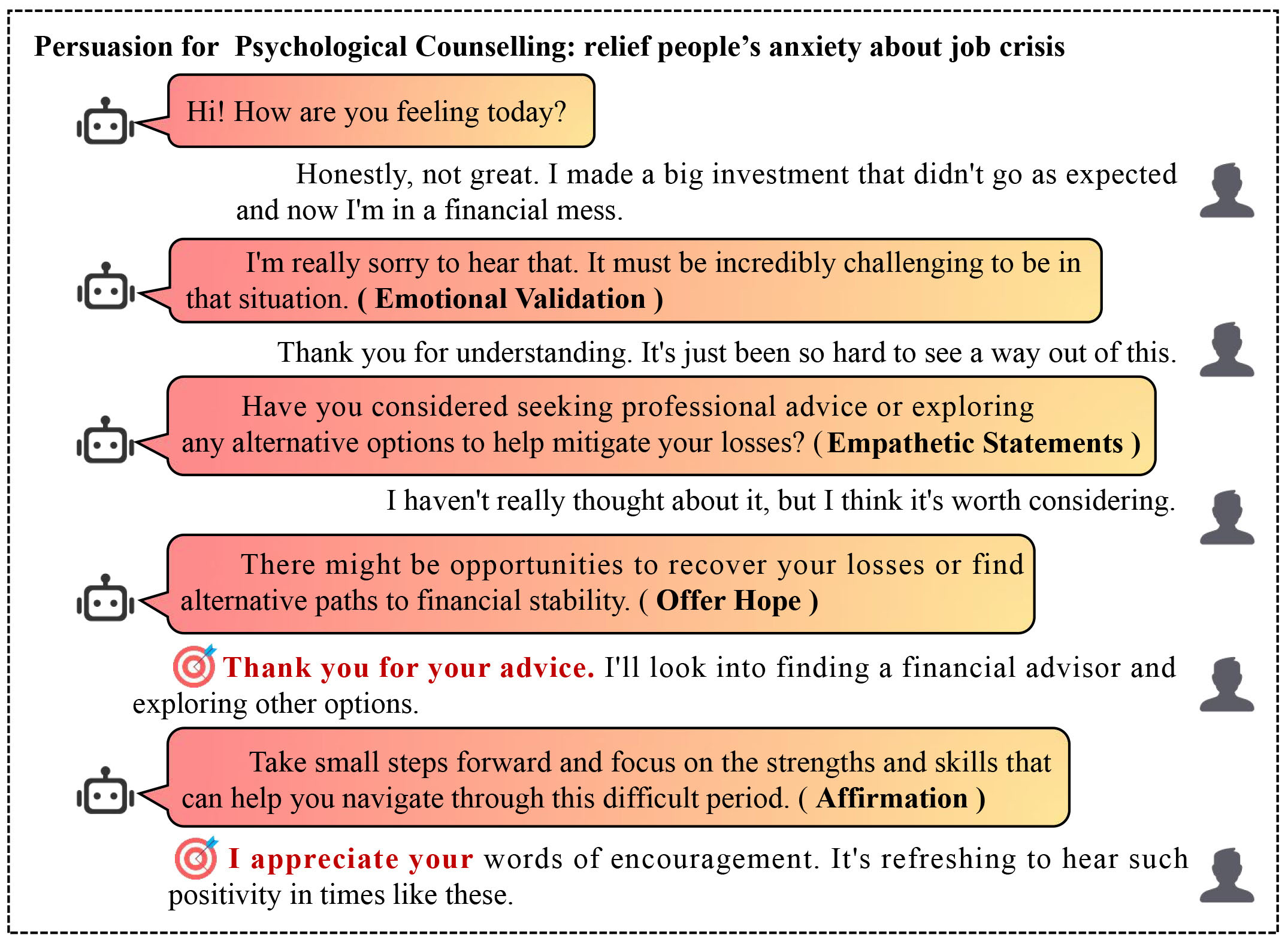}
\caption{The example of persuasive dialogue, where the dialogue agent persuades the user to relieve for job crisis using various persuasion strategies.}
\label{CogAgent_example}
\end{figure}

CogAgent aims to integrate a variety of cognitive strategies to ensure that the generated dialogue contents can effectively influence the persuadee in terms of their perceptions, opinions, or attitudes \cite{dijkstra2008psychology, wang2019persuasion}.
CogAgent has great potential in many scenarios, such as counseling depressed children \cite{kolenik2021intelligent}, persuasion for social good \cite{wang2019persuasion}, winning debates \cite{slonim2021autonomous}, and recommending items to users \cite{zhou2020towards, kang2020recommendation}. 
Fig.~\ref{CogAgent_example} depicts a persuasive dialogue example. The dialogue agent persuades the user to reduce anxiety from job crises using various persuasion strategies. 
The social and communicative dynamics behind persuasive dialogue contexts are complex. Effective and successful persuasive dialogue does not mechanically convey target viewpoints to persuadees but rather empathetically addresses persuadees through social and emotional communications \cite{chen2022seamlessly}. Thus, persuasive dialogues are not strictly task-oriented but are carried around tasks with additional cognitive strategies to build trust and empathy with persuadees, leading to a smooth persuasive process. 

As an emerging research area, an in-depth survey of the existing academic efforts is necessary. 
Duerr \textit{et al.} \cite{duerr2021persuasive} broadly reviews the works that use natural language generation to automatically detect and generate persuasive texts.
Zhan \textit{et al.} \cite{zhan2022let} concentrates on the negotiation dialogue system, a typical type of persuasive dialogue system, and comprehensively summarizes benchmarks, evaluations, and methodologies of negotiation dialogue systems. 
De\-ng \textit{et al.} \cite{deng2023survey} provide an overview of the prominent problems and advanced designs in proactive dialogue systems, which treats persuasive dialogue as the subset of the proactive dialogue.
Compared with these surveys, we provide a comprehensive review of concepts, challenges, methodologies, and applications in the field of cognitive strategy-enhan\-ced persuasive dialogue. We formalize the definition of cognitive strategies extended from cognitive psychology theory. Based on the formalized concept model and generic system architecture, we summarize representative research in the field of CogAgent from a systematic perspective. Furthermore, benchmarks, evaluation metrics, and thoughts on promising research trends are analyzed to promote the research progress. To sum up, our contributions are summarized as follows.

\begin{itemize}
    \item Drawing from cognitive psychology theories, we formalize the definition of cognitive strat\-egies, and present the concept model and gen\-eric system architecture of CogAgent, to provide an overall picture for the summary of research works.
    \item We make a profound investigation of the development in CogAgent by presenting the co\-re contributions of each work, according to the addressed challenges. Besides, we also comprehensively summarize available datase\-ts and evaluation metrics.
    \item We further discuss some open issues and pro\-mising research trends in CogAgent, including model adaptivity/generality of CogAgent, multi-party CogAgent, multimodal CogAge\-nt, etc., to promote the development of the research community.
\end{itemize}

The rest of the paper is organized as follows. In Section \ref{chap:Concept}, we first summarize the typical cognitive psychology theories and present the definition of cognitive strategies. Then we formalize the concept model of CogAgent and design a generic system architecture, followed by typical application scenarios of CogAgent.
In Section \ref{chap:works}, we first introduce the challenges faced by CogAgent and then summarize the key techniques to achieve CogAgent based on the user cognitive strategies. 
In Section \ref{chap:dataset}, we summarize the available datasets and evaluation metrics, followed by open issues and promising research trends in Section \ref{chap:open}.

\section{Formalized Concept Model and System Architecture for CogAgent}
\label{chap:Concept}
In this section, we first summarise the typical cognitive psychology theories involved in human conversations, as the theoretical foundation for the design of CogAgent.
Then we formalize the concept model for CogAgent and pres\-ent the generic system architecture to visualize the overall picture in CogAgent.

\subsection{The Cognitive Psychology Theory}
As a typical cognitive-psychological activity, the persuasion process requires the support of cognitive psychology theories to effectively model the mental changes that people experience during conversations, thus promoting the design of CogAgent. This section summarises typical cognitive psychology theories to inspire subsequent CogAgent researchers.

\subsubsection{Pre-suasion}
The concept of Pre-suasion \cite{cialdini2016pre}, proposed by the renowned authority on persuasion, Robert Cialdini, is a prominent theory in the persuasion field. 
Pre-suasion means that the success rate of persuasion can be significantly enhanced by attracting the attention of the persuadee through appropriate choic\-es of words and actions before communication or requests are conducted.
Pre-suasion emphasizes that the timing of persuasion is as important as persuasive content. 
When we intend to persuade others to accept our points, we need to consider others' perspectives and organize our conversational arguments at the appropriate time to effectively complete the persuasion process.

\subsubsection{Principle of Consistency}
The principle of consistency suggests that people usually try to maintain consistency based on what they have expressed and the commitments they have made in the past \cite{bilu2019argument}.
By planning topic paths, one can think about and define one's opinions and arguments in advance to maintain consistency and increase persuasiveness when communicating with others. 
The principle of consistency plays an important role in persuading others. 
Through consistency of statements, the persuadees will recognize that the points raised are consistent with their beliefs or opinions and will effectively increase the effectiveness of persuasion.

\subsubsection{Theory of Mind}
The theory of mind (ToM) \cite{premack1978does} suggests that effective questions and answers in communications are based on a shared world of experiences and referents between interlocutors.
To communicate effectively, people model both the mental states of their listeners and the effects of their behavior on the world, and then react to and predict the behavior of others. This ability to understand and infer human intentions is defined as a ToM.
One way to imitate ToM is to observe others' perspectives in various situations and to derive a set of rules that affect their perspectives and emotions. When the same or highly similar scenarios reoccur, we can make reasonable behavioral or emotional predictions accordingly. Many researchers explicitly model ToM as a concrete cognitive process to ensure that dialogue agents can access potential human psychological states and cognitive processes \cite{wu2023coke, sap2022neural, roman2020rmm}.

\begin{table*}[htbp]
  \centering
  \caption{Part of definitions and examples of persuasion strategies.}
  \scalebox{0.85}{
    \begin{tabular}{|m{1.8cm}<{\centering}|m{6.5cm}<{\centering}|m{11cm}<{\centering}|}
    \hline
    \textbf{Strategy } & \textbf{Definition} & \textbf{Example} \\
    \hline
    \textbf{Present of Facts} & Using factual evidence (e.g., official news reports, statistics) and a credible reasoning process to persuade others & In recent months, the demand for residential properties has become extremely high. The price of residential property has risen almost twenty percent. \\
    \hline
    \textbf{Challenges and Inquiries} & Expressing disbelief or opposition to the other side's viewpoints and providing strong rebuttal evidence to enhance persuasiveness & Really? I don't agree. This Star Wars episode was incredible! \\
    \hline
    \textbf{Emotional Resonance} & \multirow{3}[10]{=}{Eliciting specific emotions to influence others' attitudes} & State-of-the-art special effects are the main reason for the success of previous episodes, so audiences have high expectations for this one, and I don't think they will be disappointed \\
\cline{1-1}\cline{3-3}    \textbf{Eliciting Anger} & \multicolumn{1}{c|}{} & If that's the case, there's not much point in further discussion. We might as well call the whole deal off. \\
\cline{1-1}\cline{3-3}    \textbf{Eliciting Guilt} & \multicolumn{1}{c|}{} & Come on, you can at least try a little, besides your cigarette. \\
    \hline
    \textbf{Self-modeling} & Indicating one's intention to
act and choosing to act as a role model for the
persuadee to follow & That still leaves a gap of 20 dollars to be covered. Let's meet each other halfway once more, then the gap will be closed and our business completed. \\
    \hline
    \textbf{Building Trust} & Building rapport and psychological trust through a harmonious conversation & I'm glad we've agreed on price. We'll go on to the other terms and conditions at our next meeting. \\
    \hline
    \textbf{Courtesy Tips} & Expressing gratitude, approval, praise, etc. to lower the other party's psychological defenses & I know exactly what you mean. Hearing that song gives me a nostalgic feeling. \\
    \hline
    \textbf{Compromise} & Expressing concessions on time to avoid being too intense in the guidance process and causing the other party to end the conversation & I think it unwise for either of us to insist on his price. How about meeting each other halfway so that business can be concluded? \\
    \hline
    \textbf{Attachment of Views} & Expressing kindness and concern through active listening and to some extent seconding the other person's point of view & Better late than never. \\
    \hline
    \end{tabular}%
    }
  \label{strategy_def1}%
\end{table*}%

\subsubsection{Rhetoric}
Aristotle, one of the earliest masters of the art of persuasion, proposes three basic elements of persuasion: ethos (credibility), pathos (emotions), and logos (logic) in his work, The Philosophy of Rheto\-ric \cite{campbell1988philosophy}.
These principles serve as a guide to effective persuasive communication. By establishing credibility, appealing to emotions, and applying logical reasoning, one can effectively persuade others to accept his propositions. Aristotle's insights in Rhetoric remain highly influential not only in the field of persuasive dialogue but also in shaping our understanding of aesthetics and related concepts.

Credibility represents the identification of persuaders, including their identity and moral character, which influences the persuasiveness of the speaker.
Aristotle in his Rhetoric explains in detail the three elements that affect credibility, namely wisdom, virtue, and goodwill. Wisdom includes elements such as breadth of knowledge, expertise, and authority. Virtue includes elements such as fairness, honesty, and dignity.
By demonstrating wisdom, virtue, and goodwill, persuaders can enhance their persuasiveness and foster the trust and reliability of persuadees.
Combining essential elements of credibility can greatly enhance the effectiveness of CogAgent.
Emotion refers to the expression of sentiments during the persuasion process, thus lowering people's psychological defenses in accepting persuasive content. Aristotle stated that we cannot persuade others through rationality, but can achieve it with emotion. Emotional expressions play an important role in changing the cognitive decisions of others. The use of emotionally charged content and expressions can be more effective in eliciting agreement and empathy from the persuadee.
Logic refers to the use of inherent factual logic, causality, or other rational factors in expressions to gain persons' trust and persuade them to change their perceptions. By presenting coherent logical arguments, supported by factual data and authoritative sources, persuaders can establish credibility, gain persuadees' perceptions, and change their opinions.

The cognitive psychology theories, which can be used to model the dynamics of human cognitive psychological status, provide a solid foundation for CogAgent.
Under the guidance of cognitive psychology, we can comprehensively investigate and model explicit cognitive factors and strategies that can change users' cognitive psychological states, such as logical expressions and emotional appeals. These cognitive strategies can facilitate CogAgent to understand the psychological state of the persuadee and enhance the persuasiveness of responses from multiple perspectives to achieve mo\-re efficient persuasion processes.

\subsection{Cognitive Strategy}
To achieve efficient persuasion, it is necessary to integrate various cognitive strategies for precisely responding to the psychological changes of the persuadee. Evolved from cognitive psychological theories, we categorize cognitive strategies into three aspects, \textbf{persuasion strategy}, \textbf{topic path planning strategy}, and \textbf{argument structure prediction stra\-tegy}, detailed as follows.

\subsubsection{Persuasion Strategy} 
The persuasion strategy aims to influence or change the perceptions, opinions, attitudes, or behaviors of persuadees from a psychological standpoint, thro\-ugh the use of linguistic techniques of expression, such as logical appeal, foot-in-the-door, and self-disclosure \cite{fogg2002persuasive, wang2019persuasion, liu2021towards}. 
Based on existing research, we construct a comprehensive and effective set of persuasion strategies that can achieve persuasive goals, inspired by the theory of mind, the rhetoric, and other psychology theories. We formalize the definitions and examples of expressions of persuasion strategies, as shown in Table~\ref{strategy_def1} and Table~\ref{strategy_def2}.

\begin{table*}[htbp]
  \centering
  \caption{Part of definitions and examples of persuasion strategies.}
  \scalebox{0.85}{
    \begin{tabular}{|m{2.2cm}<{\centering}|m{7.8cm}<{\centering}|m{9.5cm}<{\centering}|}
    \hline
    \textbf{Strategy} & \textbf{Definition} & \textbf{Example} \\
    \hline
    \textbf{Problem Decomposition} & Decomposing the ultimate persuasion goal into sub-issues and stepping through the persuasion process & Let me get down some information about your apartment first. what is your property's address? \\
    \hline
    \textbf{Social Identity} & Gaining psychological support from the other person by emphasizing group and identity belonging & I know. I have been a subscriber for the past two years. \\
    \hline
    \textbf{Herd Mentality} & Presenting a viewpoint that is recognized or accepted by the majority of people and persuading the other side to accept it & There was always a good round of applause every time she sang. \\
    \hline
    \textbf{Expression of Disgust} & \multirow{3}[8]{=}{Expressing a particular point of view or emotion to emphasize the persuasive content} & Oh, my god! I look so old. I look as if I were 40. I think it's time for some plastic surgeries. \\
\cline{1-1}\cline{3-3}    \textbf{Expression of Empathy} & \multicolumn{1}{c|}{} & I know, dear. I am too. But we've just been too busy to look for a house. \\
\cline{1-1}\cline{3-3}    \textbf{Expression of Views} & \multicolumn{1}{c|}{} & That means the apartment has furniture in it.  \\
    \hline
    \textbf{Logical Appeal} & Enhancing the credibility of persuasive content through the logical and reasoning process & It certainly is. But to tell you the truth, the room is so large that I can share it with someone else, and that will decrease the total amount of the rent. \\
    \hline
    \textbf{Task Inquiry} & Asking questions related to persuasive goals & That might be going overboard a bit. How about just that scarf with a bracelet? \\
    \hline
    \textbf{Personal Story} & Using narrative examples to illustrate the positive outcomes of your actions to inspire others to follow suit & Yes, I'm sure I've done a lot of house painting in my life. If I got even a tiny drop of paint on her furniture, she would get furious. So I learned to be very picky. \\
    \hline
    \textbf{Refutation of Objections} & Directly refuting the other side's point of view & Not necessary. If we use a realtor to find a house, it will be more expensive. \\
    \hline
    \textbf{Greeting} & Greeting at the beginning of a dialogue & Hi there! How are you doing today? \\
    \hline
    \end{tabular}%
    }
  \label{strategy_def2}%
\end{table*}%

Numerous studies \cite{he2018decoupling, wang2019persuasion, joshi2020dialograph, cheng2022improving} have demonstrated that persuasion strategies can effectively enhance the persuasiveness of the dialogue content. 
How to reasonably select the appropriate strategies according to the dialogue context and the perusadee's psychological state to generate a persuasive dialogue response is crucial to achieving high-quality CogAgent.

\subsubsection{Topic Path Planning Strategy}
The topic path planning strategy aims to plan the topic transition sequence during the persuasive dialogue process, to ensure the dialogue coherence and the progress of the dialogue towards the persuasive target. The persuasive dialogue agent shou\-ld smoothly navigate between topics to reduce irrelevant associations of the persuadee and the difficulty of the persuasion process \cite{cacioppo1979effects, cialdini2007influence}. 
The topic path planning strategy is widely employed in target-guid\-ed persuasive dialogue systems \cite{ni2022hitkg, tang2023eagle}. 
Starting from the topic of interest to the persuadee, the persuasive dialogue agent needs to gradually and smoothly transfer the conversation topic to the persuasive target to improve the persuadee's psychological acceptability and ensure the persuasive effect. How to plan the reasonable topic path and generate an in-depth multi-turn persuasive conversation according to the corresponding topics is to be explored.

\subsubsection{Argument Structure Prediction Strategy}
Argument structure prediction strategy is designe\-d to predict persuasive and authoritative argument su\-rrounding the discussed topic, thereby enhancing the credibility of persuasive dialogue contents and convincing the persuadee of the plausibility of the proposed claims \cite{petty2012communication, swanson2015argument, chakrabarty2019ampersand}. Persuasive dialogue age\-nts need to be equipped with a large-scale library of arguments and counter-arguments. By predicting reasonable argument structures based on specific persuasive topics, dialogue agents can incorporate coherent argumentation skills, such as citing authorities and providing convincing arguments and evidence, to effectively enhance the plausibility of dialogue contents and the credibility of the persuasion process. 
The argument structure prediction strategy has been extensively explored in the field of debate dialogue, where debaters often consider argument structures to express viewpoints with clarity, logical coherence, and compelling evidence \cite{rach2021argument, slonim2021autonomous, wambsganss2021arguetutor}. With the argumentative structure, the whole persuasive process can be progressed incrementally, and the overall organization, logical coherence, and credibility of the persuasive process can be significantly increased. How to mine the supporting argument structures based on the dialogue context and reasonably integrate the argument structures into dialogue contents to enhance the credibility of persuasive dialogue is to be investigated.

\subsection{Formalized Concept Model for CogAgent}
Based on the definitions of cognitive strategies, we define the dialogue system that is incorporated with cognitive strategies to accomplish persuasive tasks through smooth and accessible conversations as C\-ognitive Strategy-enhanced Persuasive Dialogue (\-\textbf{CogAgent}). We introduce the formalized concept mo\-del of CogAgent as follows.

Typically, given the dialogue context sequence $H=\{(Q_{1},A_{1}),...,(Q_{S-1},A_{S-1})\}$ with $S$-1 turns, wh\-ere $Q_{i}$ and $A_{i}$ are the dialogue query and response at the $i$-th dialogue turn, and the current dialogue query $Q_{S}=(q_{1},...,q_{m})$ with $m$ words, the objective of general dialogue system is to generate the dialogue response $A_{S}=(a_{1},...,a_{n})$ with $n$ words.
The modern dialogue systems usually follow the encoder-decoder architecture \cite{huang2020challenges,ni2023recent} or decoder-on\-ly architecture \cite{touvron2023llama, bubeck2023sparks}. For the encoder-decoder architecture, the encoder aims to transform input text sequence into vector representations using LSTM \cite{hochreiter1997long}, Transformer \cite{vaswani2017attention} or other advanced neural models, as shown in Eq.~\ref{Encoder}.

\begin{equation}
\mathbf{Q_{S}}, \mathbf{H} = \textbf{Encoder}(Q_{S}, H)
\label{Encoder}
\end{equation}

Based on the semantic vectors of dialogue context and input query, the decoder generates the dialogue response word by word in an auto-regressive manner, as shown in Eq.~\ref{Decoder}, where $a_{t}$ is the $t$-th words in the response.

\begin{equation}
P(A|\mathbf{Q_{S}},\mathbf{H}) = \prod_{t=1}^{n} p(a_{t}|\mathbf{Q_{S}},\mathbf{H},a_{<t})
\label{Decoder}
\end{equation}

For the decoder-only architecture, all input text sequences will concatenated into a uniform sequen\-ce with special tokens, and then the decoder also generates the response in a word-by-word manner.
The general dialogue system can generate smooth and fluent responses based on the dialogue context. To generate persuasive dialogue content, it is essential to combine three kinds of cognitive strategies.

\begin{figure*}[t]
\centering
\includegraphics[scale=0.6]{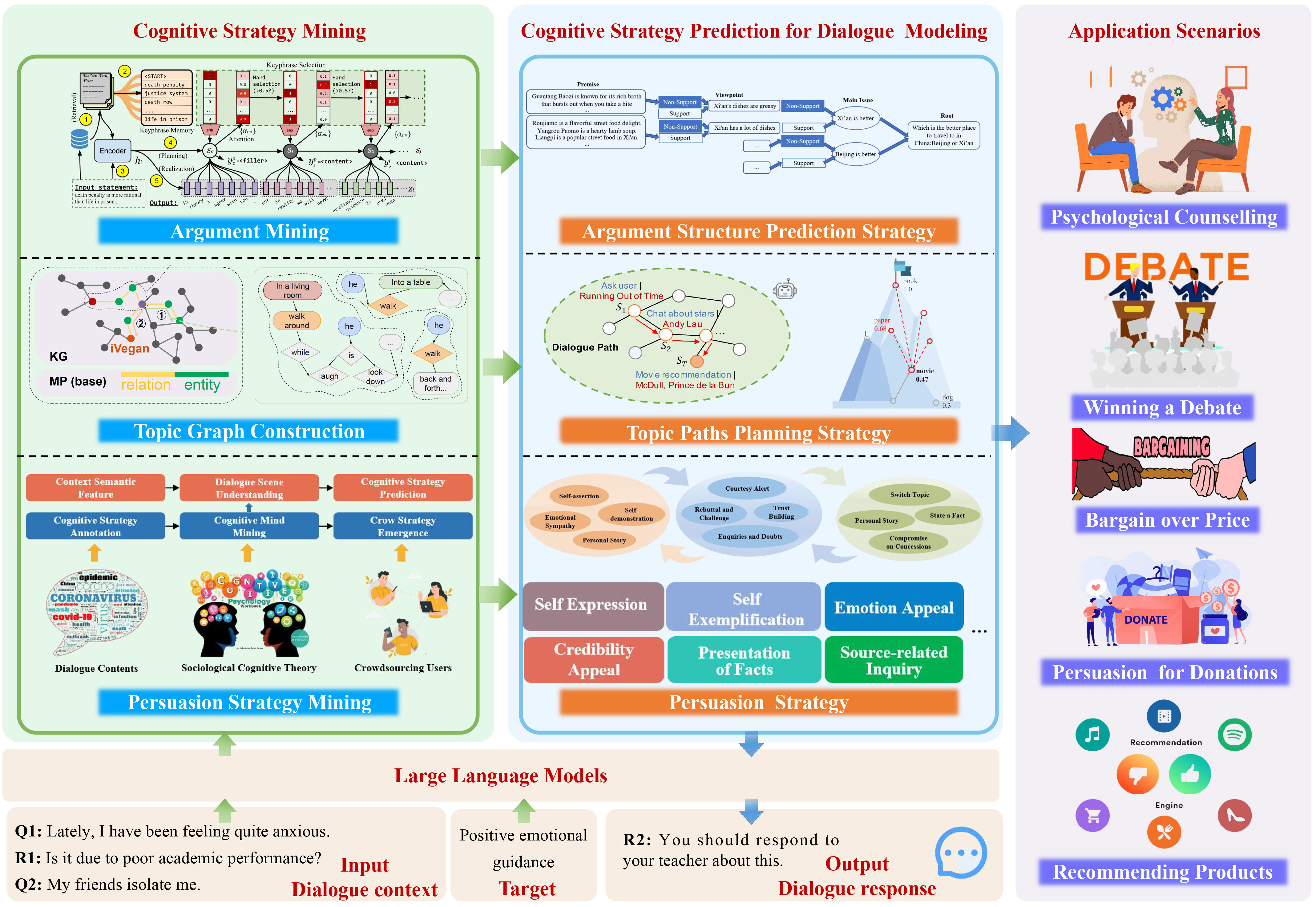}
\caption{The generic system architecture of CogAgent.}
\label{System_architecture}
\end{figure*}

Based on the definitions of three cognitive strategies, we give the formalized definition of CogAgent. Gi\-ven the dialogue context and the current query, CogAgent needs to first predict the persuasion strategy $Per$, conversation topic $Top$, and the argument content $Arg$ based on the current dialogue content, as follows.

\begin{equation}
Per, Top, Arg = Str\_Pre(\mathbf{Q_{S}}, \mathbf{H})
\label{Strategy_pre}
\end{equation}
where $Str\_Pre$ refers to the cognitive strategy predictor. Then the dialogue decoder generates the dialogue response word by word conditioned on additional cognitive strategies, as shown in Eq.~\ref{CogAgent_form}.

\begin{equation}
\begin{split}
P(A|\mathbf{Q_{S}},\mathbf{H},Per,Top,Arg) = \\
\prod_{t=1}^{n}p(a_{t}|\mathbf{Q_{S}},\mathbf{H},Per,Top,Arg,a_{<t})
\label{CogAgent_form}
\end{split}
\end{equation}

\subsection{Generic System Architecture}
After the concept model of CogAgent, we present the generic system architecture of CogAgent, as shown in Fig~\ref{System_architecture}. The overall process of CogAgent starts from the semantic understanding of dialogue context and the persuasive target, powered by LLMs (e.g., ChatGPT, LLaMa, Claude, ChatGLM). The input text will be encoded into semantic embeddings for subsequent processes. 
The \textit{Cognitive Stra\-tegy Mining} part 
is responsible for mining cognitive strategies, including persuasion strategies, topic paths over knowledge graph, and argument structure of topics. The \textit{Cognitive Strategy Prediction for Dialogue Modelling} part predicts appropriate cognitive strategies based on dialogue context and enhances the linguistic expression, logical structure, and other persuasive aspects of responses.

The persuasion strategy mining process first min\-es various kinds of persuasion strategies through crow\-d strategy emergence based on cognitive psychology theories. According to the dialogue context, the persuasion strategies to be used in subsequent roun\-ds of dialogue will be predicted.
The topic graph construction process constructs topic graphs or topic paths and then plans the wandering paths of topics for persuasion according to the dialogue context and persuasion strategies.
The argument mining process first constructs a complete argument structure from credible data sources and then predicts the arguments needed for persuasion based on the above cognitive strategies.
Finally, the cognitive strategies-enhanced dialogue context will be fed into LLMs to generate persuasive dialogue responses, for numerous applications, such as psychological counseling, bargaining, and persuasion for social good.

\subsection{Application Scenarios}
Persuasive dialogue system has widespread applications in daily life. It is an ongoing effort of the academic/industry researchers to conduct persuasive dialogue with users to achieve persuasive targets, summarized as follows.

\textbf{Persuasion for social good.} 
Persuasion for social good is a typical persuasive dialogue scenario where people are persuaded to donate money or goods to charities for social good purposes, such as children's aid and natural disaster relief. Many researchers have explored how to combine persuasion strategies to promote users' donation behavior.
For example, Wang \textit{et al.} \cite{wang2019persuasion} provide an insightful analysis of what persuasion strategies are effective for what types of personal characteristics of users.
Mishra \textit{et al.} \cite{mishra2022pepds} propose a Reinforcement Learning (RL) based persuasive dialogue system with an efficient reward function consisting of five different sub rewards, Persuasion, Emotion, Politeness-Strategy Consistency, Dialogue-Coherence, and N\-on-repetitiveness.
Chen \textit{et al.} \cite{chen2022seamlessly} produce a modular persuasive dialogue system that seamlessly integrates factual information and persuasive content into generated dialogue response using the conditional language model.

\textbf{Persuasion for psychological counseling.}
The frequent occurrence of mental diseases, such as depression, makes mental health gradually receive extensive attention from society \cite{walker2015mortality, xu2022survey, liang2023identifying}. Psychological counseling aims at reducing people’s emotional distress and helping them understand and wo\-rk through the challenges that they face. Relieving the psychological pressure of the persuaded through conversation holds profound significance for the pe\-rsuasive dialogue system.
Extensive studies have explored the possibility of using persuasive dialogue systems to provide psychological counseling.
For example, Liu \textit{et al.} \cite{liu2021towards} collect the Emotion Support Conversation dataset (ESConv) with well-desi\-gned persuasion strategy annotation to train dialogue system to provide emotional support through dialogue interactions.
Zhou \textit{et al.} \cite{zhou2022case} build a commonsense cognition graph and an emotional concept graph based on commonsense knowledge from COMET \cite{bosselut2019comet} and concept knowledge from ConceptNet \cite{speer2017conceptnet}. The two kinds of knowledge are alig\-ned to generate dialogue responses for emotional support.

\textbf{Persuasion for negotiation.}
Negotiation is a common real-life persuasion scenario in which two parties negotiate through ongoing conversations to persuade the other party to accept the terms or demands they make to maximize their interests. Negotiation is a necessary means of facilitating agreements among people and improving the efficiency of society. There have been several studies using persuasive dialogue systems to achieve negotiation.
For instance, Joshi \textit{et al.} propose DIALOGRAPH \cite{joshi2021dialograph}, a negotiation dialogue system that explicitly incorporates dependencies between sequences of strategies into graph neural networks.
Nortio \textit{et al.} \cite{nortio2022fear} embark on an exploration of persuasive techniques in international negotiations, emphasizing the significance of persuasive strategies during the negotiation process.

\textbf{Persuasion for debate.}
Debate is a professional persuasive scenario in which debaters persuade the opponent and the audience to accept their viewpoints by planning their arguments wisely and arguing their points from multiple perspectives.
Many researchers have explored the automatic generation of persuasive arguments from online discussions or debate competitions \cite{sakai2020hierarchical, rach2020increasing, rach2021argument}.
Slonim \textit{et al.} \cite{slonim2021autonomous} introduce Project Debater, an autonomous debating system that can engage in a competitive debate with humans.

\textbf{Persuasion for recommendation.} Engaging in dialogue-based recommendations for movies, products, and other such aspects proves to be a highly practical application of a persuasive dialogue system. 
To achieve successful recommendations, it is crucial to employ a persuasion strategy to facilitate rapid user comprehension and acceptance of the recommendations.
For example, Gupta \textit{et al.} \cite{gupta2022target} propose to decompose the recommendation response generation process into first generating explicit commonsense paths between the source and persuasive target followed by generating responses conditioned on the generated paths.

\section{Research Challenges and Key Techniques}
\label{chap:works}
Due to the complexity of modeling the psychological changes in the persuasive conversation, many critical challenges in CogAgent need to be address\-ed. In this section, we first detail these challenges faced by CogAgent, and then conduct a comprehensive investigation of representative works of CogAgent according to the adopted cognitive strategies, i.e., the persuasion strategy, the topic path planning strategy, and the argument structure prediction strategy.

\subsection{Research Challenges in CogAgent}
\textbf{Exhaustive mining of cognitive strategies}.
Ps\-ychology defines human cognition as the process by which a person encounters, perceives, and understands things \cite{mondal2022unifying, shettleworth2009cognition}. The formation and evolution of human cognition is an extremely complex process involving knowledge, personality, emotion, an\-d many other aspects.  
Effective persuasive dialogue changes people's feelings and perceptions about thi\-ngs through persuasive strategies that convince people to change their opinions and behaviors \cite{wang2019persuasion, nguyen2008designing}.
The\-refore, it is a great challenge to build a complete set of cognitive strategies from the perspective of cognitive psychology by mining cognitive strategies that can effectively change the way human beings perceive and understand things.
Several researchers have defined some persuasion strat\-egies based on cognitive psychology theories (e.g., \textit{logical appeal} and \textit{emotion appeal} from \cite{wang2019persuasion}, \textit{self-disclosure} from \cite{liu2021towards}). However, most of these strat\-egies are task-specific and not exhaustive enough to cope with generalized persuasion scenarios. 
How to construct well-defined cognitive strategies from multiple perspectives needs to be explored in depth.

\textbf{Modeling and selecting of cognitive strategies.} 
In persuasive dialogues, people usually dynamicall\-y choose different persuasive strategies depending on different persuasive goals and the evolving conversational contexts. 
Persuasive strategies contain complex semantic patterns, rather than mere names or descriptions \cite{orji2017persuasive, ham2011making}. 
How to model the implicit associations between strategy definitions and linguistic expressions, and precisely select cognitive strategies according to the dialog context to facilitate the smooth flow of the persuasive dialog process is a serious challenge.
Some researches have explored how to select appropriate cognitive strategies based on the dialogue context \cite{joshi2021dialograph, cheng2022improving}. 
The appropriate selection of cognitive strategies is a critical step for CogAgent to simulate humans in persuasive conversations and is essential for achieving high-quality persuasive conversations.

\textbf{Integrating cognitive strategies into models.} 
As defined at the cognitive psychology level, cognitive strategies are more abstract semantic concepts. Data-driven neural network models (DNNs), even LLMs, remain superficial in the understanding of cognitive strategies. 
How to facilitate DNNs to learn the profound semantics of cognitive strategies, to rationally integrate cognitive strategies into the generation of persuasive dialogues, and to improve the persuasiveness of CogAgent, is quite cha\-llenging.
Graph-based, \cite{joshi2021dialograph}, reinforcement lear\-ning-based \cite{samad2022empathetic} and other advanced methods are investigated to integrate cognitive strategies into persuasive dialog generation.
It is promising to integrate cognitive strategies with the outstanding language comprehension ability of LLMs.

\textbf{Absence of evaluation metrics.}
To improve the quality of persuasive dialogue, the performance of CogAgent needs to be evaluated accurately and co\-mprehensively.
However, existing evaluation metrics for dialog systems (e.g., BLEU \cite{papineni2002bleu}, METEOR \cite{banerjee2005meteor}, ROUGE-L \cite{lin2003automatic}) are usually evaluated at the level of word similarity or semantic similarity between generated responses and ground truth, without taking into account the effectiveness of persuasive strategies, the rationality of persuasive path planning, and the richness of argument structure.
It is a challenge to develop comprehensive and reasonable evaluation metrics to accurately evaluate the quality of CogAgent, incorporating the characteristics of persuasive dialog systems.

\subsection{Persuasion Strategy-based CogAgent}
Incorporating persuasion strategies to enhance the persuasiveness of dialog responses is an important research direction in CogAgent. By using specific persuasive strategies, CogAgent can express the pe\-rsuasive content in a way that is more acceptable to the persuadees, thus accomplishing the persuasive goals more smoothly.
As abstract psychological concepts, how to select appropriate persuasion strategies according to the dialogue context and gui\-de the generation of responses is an important research question. In this section, we make an investigation of the employment of persuasion strategies in CogAgent, summarized in Table~\ref{tab:Persuasion_CogAgent}.

\begin{table*}[htbp]
  \centering
  \caption{Representative works of persuasion strategy-based CogAgent.}
  \scalebox{0.85}{
    \begin{tabular}{|m{5cm}<{\centering}|m{3cm}<{\centering}|m{11cm}<{\centering}|}
    \hline
    Solution & Work  & Description \\
    \hline
    \multirow{2}[4]{*}{\shortstack{Strategy classifying based \\ on dialogue context}} & Wang et al. \cite{wang2019persuasion} & Proposing a classifier to predict persuasion strategies in dialogue using context and sentence features. \\
\cline{2-3}          & He et al. \cite{he2018decoupling} & Decoupling strategy selection and response generation in CogAgent for predicting strategy and generating responses based on dialogue history. \\
    \hline
    \multirow{2}[4]{*}{Persuasion strategy planning} & Cheng et al. \cite{cheng2022improving} & Proposing lookahead heuristics to estimate future user feedback after using the specific strategy. \\
\cline{2-3}          & Yu et al. \cite{yu2023prompt} & Using Monte Carlo Tree Search for persuasion strategy planning without model training. \\
    \hline
    \multirow{2}[4]{*}{Graph-based strategy incorporation} & Joshi et al. \cite{joshi2020dialograph} & Using GNNs to model strategies, dialogue acts, and dependencies in graph structures for response generation. \\
\cline{2-3}          & Zhou et al. \cite{zhou2019augmenting} & Modeling both dialogue context semantic and persuasion strategy history with finite state transducers. \\
    \hline
    \multirow{2}[4]{*}{\shortstack{Knowledge-enhanced \\ strategy modeling}} & Jia et al. \cite{jia2023knowledge} & Introducing a knowledge-enriched encoder and memory-enhanced strategy module for dynamic emotion and semantic pattern modeling. \\
\cline{2-3}          & Chen et al. \cite{chen2022seamlessly} & Designing RAP for dynamic factual and persuasive responses based on knowledge and individual persuasion strategies. \\
    \hline
    \multirow{2}[4]{*}{Novel integration mechanism} & Mishra et al. \cite{mishra2022pepds} & Creating an RL reward function to enhance consistency in politeness strategy, persuasiveness, and emotion acknowledgment in persuasive dialogue. \\
\cline{2-3}          & Tu et al. \cite{tu2022misc} & Proposing a novel model MISC, which firstly infers the user’s fine-grained emotional status, and then responds skillfully using a mixture of strategies. \\
    \hline
    \end{tabular}%
    }
  \label{tab:Persuasion_CogAgent}%
\end{table*}%

\subsubsection{Strategy Classification based on Dialogue Context}
A straightforward approach to fusing persuasion strategies in persuasive conversations is to predict a strategy label (e.g., Present of Facts) based on the dialogue context and feed the strategy into the decoder with the dialogue context to generate the dialogue response.

For example, Wang \textit{et al.} \cite{wang2019persuasion} propose a persuasion strategy classifier to predict 10 persuasion strategies based on the dialogue context information and sentence-level features. The authors also analyze the impact that different people's backgrou\-nds on strategy prediction, laying the grou\-ndwork for research on personalized persuasive dialogue agents.
He \textit{et al.} \cite{he2018decoupling} decouple strategy selection and response generation in CogAgent. The dialogue manager predicts a persuasion strategy ba\-sed on the persuasion strategies in dialogue history by a sequence-to-sequence model and the response generator produces a response conditioned on the stra\-tegy and dialogue history.

\subsubsection{Persuasion Strategy Planning}
Persuasive dialogue is usually a process that lasts multiple turns, supported by successive strategies \cite{greene2003handbook, hill2009helping}. Consequently, strategy planning within a long planning horizon in CogAgent is quite important, rather than predicting a specific strategy based on the dialogue history.
Several studies focus on long-term planning of persuasion strategies, making CogAgent more efficient in reaching persuasion goals.

For instance, Cheng \textit{et al.} \cite{cheng2022improving} firstly adopt an A* search algorithm for persuasion strategy planning. When predicting the appropriate strategy in each dialogue turn, look-ahead heuristics are proposed to estimate future user feedback after using the specific strategy, thus considering the long-term effect of persuasion strategies.
The proposed lookahead method requires abundant annotated data, affecting the application to broader persuasive dialogue scenarios.
To overcome this bottleneck, Yu \textit{et al.} \cite{yu2023prompt} prompts LLMs to perform persuasion strategy planning by simulating future dialogue interactions using the Monte Carlo Tree Search (MC\-TS) algorithm. This method requires no model trai\-ning and can therefore be adapted to any persuasion scenario.


\begin{figure*}[htbp]
    \centering
    \includegraphics[width=.65\textwidth]{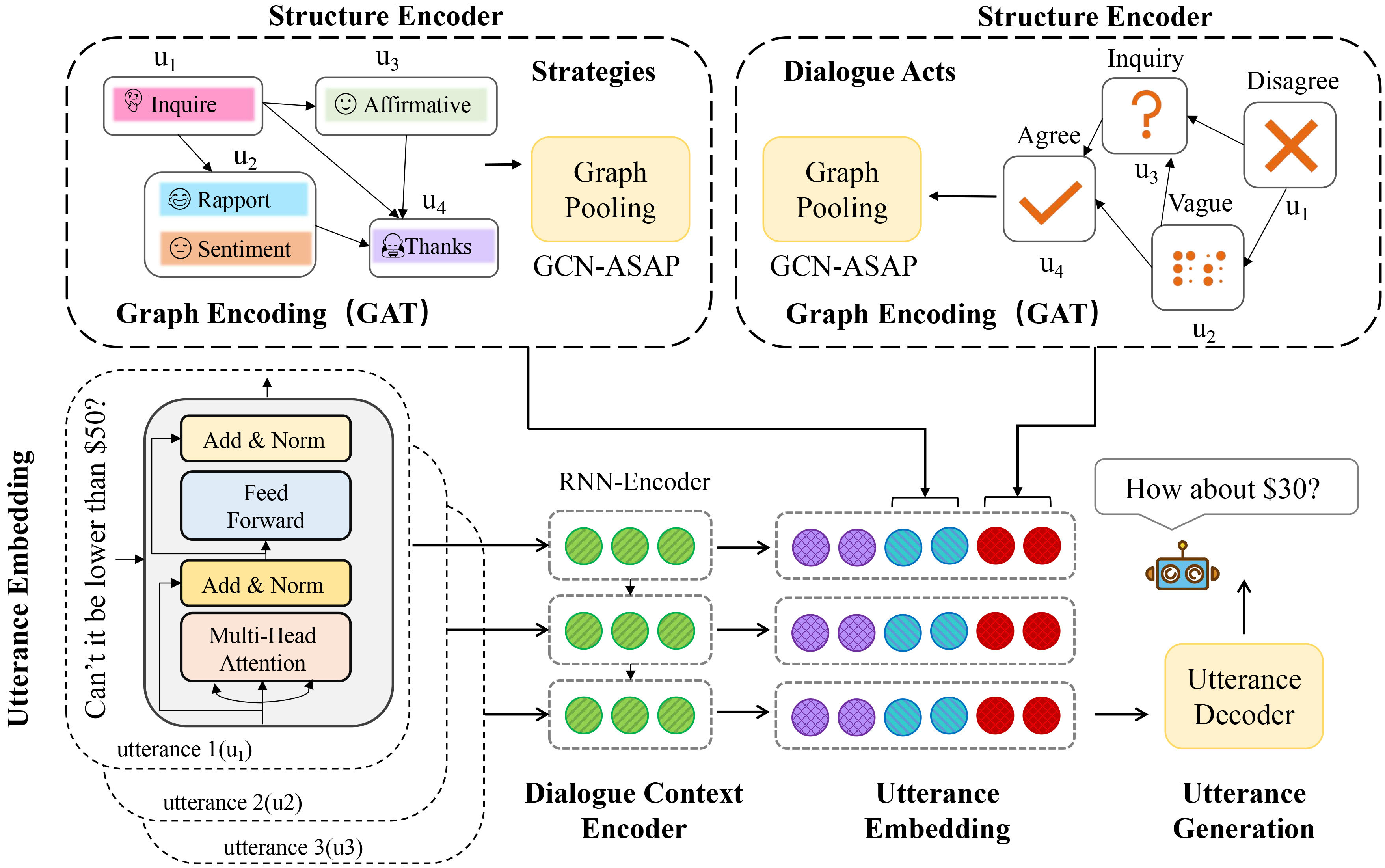}
    \caption{Overview architecture of DIALOGRAPH which models persuasion strategies as graph structure.}
    \label{DIALOGRAPH}
\end{figure*}

\subsubsection{Graph-based Strategy Incorporation}
Graph Neural Networks (GNNs) \cite{kipf2016semi, velivckovic2018graph, wu2020comprehensive} can combine the benefits of interpretability and expressivity, benefiting from encoding graph-structured data through message propagation. Due to the human brain's reasoning process to capture semantic associations, graph-based methods have been widely used in various tasks \cite{wu2023graph,wang2023towards}. 
Numerous researche\-rs have embarked on exploring the potential of gra\-ph-based methods for incorporating persuasion str\-ategies in CogAgent.

For example, Joshi \textit{et al.} \cite{joshi2020dialograph} introduce DIAL\-OGRAPH, as shown in Fig~\ref{DIALOGRAPH}, a persuasive dialogue system that incorporates persuasion strategies and dialogue acts using GNNs. DIALOGRAPH models persuasion strategies in multi-turn dialogue context and their dependencies as graph structures and incorporating strategies into response generation using hierarchical graph pooling-based approaches.
Zhou \textit{et al.} \cite{zhou2019augmenting} propose to model both dialogue context semantic and persuasion strategy history finite state transducers (FSTs). 
To model the persuasion factors affecting the persuasive content of dialogues, Liu \textit{et al.} \cite{liu2022modeling} present persuasion-factor graph convolutional layers to encode and learn representations of the persuasion-aware interaction data.

\subsubsection{Knowledge-enhanced Strategy Modeling}
As concepts in cognitive psychology, persuasion strategies encompass complex semantic information and various intricate linguistic features \cite{hill2009helping, zheng2021comae}.
To comprehensively represent the complex semantics embedded within persuasion strategies, researches investigate combining external knowledge to model and mimic the the intricate patterns in strategies.

For example, Jia \textit{et al.} \cite{jia2023knowledge} propose a knowledge-enriched dialogue context encoder to model the dynamic emotion state and a memory-enhanced strategy modeling module to model the semantic patterns of persuasion strategies. The same-strategy responses are stored in the memory bank to provide more specific guidance for the strategy-
constrained response generation.
Chen \textit{et al.} \cite{chen2022seamlessly} design the
Response-Agenda Pushing Framework (R\-AP) to dynamically produce factual responses bas\-ed on knowledge facts and persuasive responses conditioned on individual persuasion strategies.

\subsubsection{Novel Integration Mechanisms}
In addition to the above studies to model and integrate persuasion strategies, researchers propose some no\-vel integration mechanisms to improve the performance of CogAgent, summarized as follows.

Combined with RL, Ya\-ng \textit{et al.} \cite{yang2021improving} propose two variants of ToM-based persuasive dialog agent, wh\-ere the explicit version that outputs the opponent type as an intermediate prediction, and an
implicit version that models the opponent type as a latent variable. Both models are optimized using reinforcement learning.
Similarly, Mishra \textit{et al.} \cite{mishra2022pepds} design an efficient reward function in RL to improve the politeness-strategy consistency, persuasiveness, and emotional acknowledgeme\-nt in persuasive dialogue.

To increase the expressed empathy and learn the gradual transition in the long response, Tu \cite{tu2022misc} introduce a MIxed Srategy-aware model (MISC) integrating COMET, a pre-trained generative commonsense reasoning
model, for emotional persuasive dialogue. The COMET knowledge tuples are adopted to enhance the fine-grained emotional understanding of users. Then MISC formulates persuasion strategy as a probability distribution over a strategy codebook to use a mixture of strategies for persuasive response generation.

To investigate the potential of LLMs in persuasive conversations, Zheng \textit{et al.} \cite{zheng2023building} first construc\-t a large-scale persuasive dialogue dataset in the emotional support domain, leveraging the generative capabilities of LLMs. Then several advanced tuning techniques (fine-tuning, adapter-tuning, Lo\-RA-tuning) are employed to 
to showcase the superiority of LLMs in persuasive dialogue generation.

\begin{table*}[htbp]
  \centering
  \caption{Representative works of topic path planning strategy-based CogAgent.}
  \scalebox{0.90}{
    \begin{tabular}{|m{3.5cm}<{\centering}|m{2.5cm}<{\centering}|m{12cm}<{\centering}|}
    \hline
    Solution & Work  & \multicolumn{1}{c|}{Description} \\
    \hline
    \multirow{3}[8]{*}{\shortstack{Reinforcement learning \\ based planning}} & Xu et al. \cite{Xu2020Knowledgegraph} & Presenting KnowHRL, a three-layer Knowledge-aware hierarchical RL-based model for coherent topic path planning and multi-turn persuasive dialogue responses. \\
\cline{2-3}          & Liu et al. \cite{liu2020GoChat} &  Hierarchical RL for conversation topic path planning, using high-level strategies and low-level responses. \\
\cline{2-3}          & Lei et al. \cite{Lei2022I-Pro} & Introducing four persuasion-related factors in the reward function to achieve persuasive goals efficiently. \\
    \hline
    \multirow{3}[8]{*}{\shortstack{Graph-based \\ planning}} & Zhong et al. \cite{zhong2021keyword} & Using commonsense knowledge graphs and GNN to enhance semantic relations between topic keywords, improving keyword-augmented response retrieval. \\
\cline{2-3}          & Zou et al. \cite{zou2021thinking} & Employing a concept graph for topic planning, utilizing an Insertion Transformer for persuasive response generation based on multi-concept paths. \\
\cline{2-3}          & Wang et al. \cite{wang2023dialogue} & Introducing a Transformer-based network for target-driven topic path planning with knowledge-target mutual attention and set-search decoding. \\
    \hline
    \multirow{2}[4]{*}{\shortstack{Novel planning \\ mechanism}} & Tang et al. \cite{tang2023eagle} & Combining various planning algorithms for robust and smooth topic path planning, incorporating a sampling strategy, flow generator, and global planner. \\
\cline{2-3}          & Wang et al. \cite{wang2021two} & Introducing a consistency-driven dialogue planning approach that utilizes stochastic processes to model the temporal evolution of the conversation path dynamically. \\
    \hline
    \end{tabular}%
    }
  \label{tab:Topic_CogAgent}%
\end{table*}%

\subsection{Topic Path Planning Strategy-based CogAgent}
In persuasive dialogues, generating engaging responses through effective topic path planning is critical to achieving persuasive targets. Topic path planning strategy is a navigation tool that enhances the coherence of the persuasion process by continuously leading users to discuss different points and topics until reaching persuasive targets. This section deepens into the intricate details of the topic path planning strategy, summarized in Table~\ref{tab:Topic_CogAgent}.

\begin{figure*}[h]
    \centering
    \includegraphics[width=.8\textwidth]{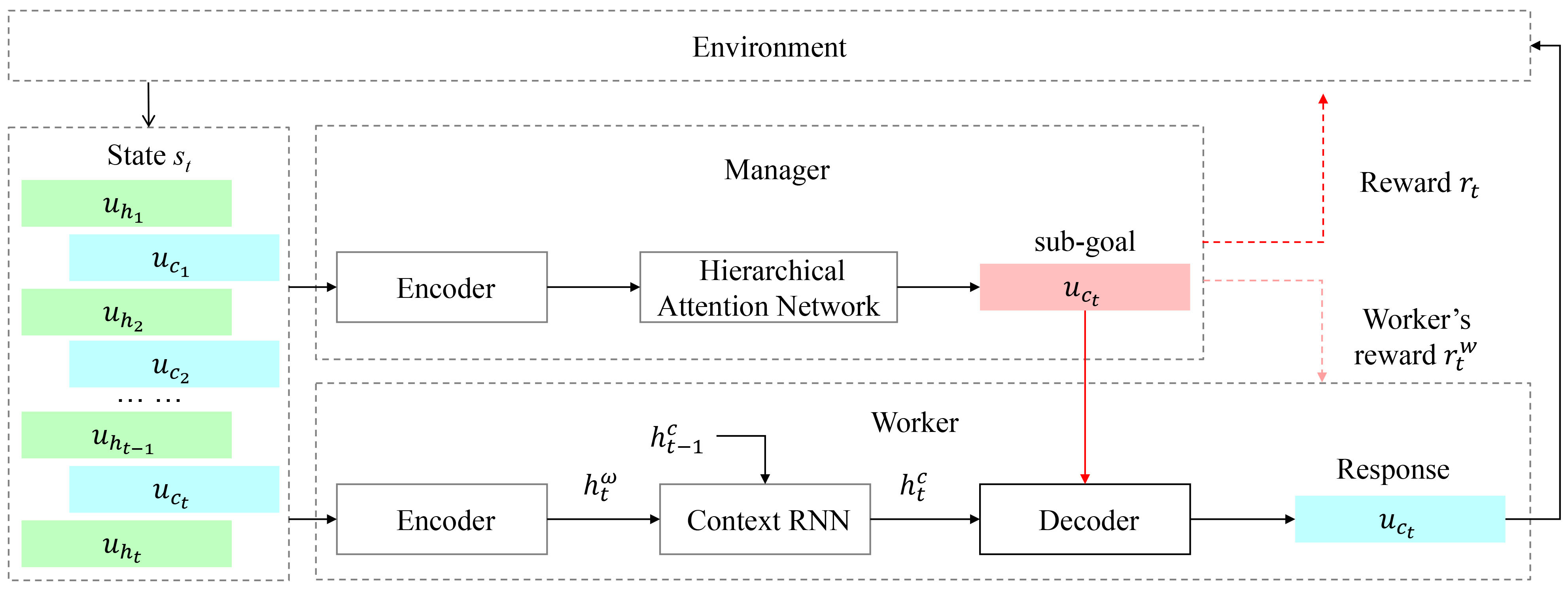}
    \caption{The overall framework of GoChat with hierarchical reinforcement learning.}
    \label{gochat-pic}
\end{figure*}

\subsubsection{Reinforcement Learning-based Planning}

In the context of topic paths planning strategy, Reinforcement Learning serves as a dynamic framew\-ork for guiding persuasive dialogue systems in a goal-oriented manner.
The core of RL is to learn the optimal sequence of actions according to the reward function and is therefore ideally suited for planning coherent topic paths in CogAgent.

For example, to achieve coherent topic path planning, Xu \textit{et al.} \cite{Xu2020Knowledgegraph} introduce a three-layer Knowledge aware hierarchical RL-based model (KnowH\-RL). The upper layer of KnowHRL plans a high-level topic sequence to track user interests toward persuasive targets. The lower layers are responsible for generating multi-turn persuasive dialogue responses.
similarly, Liu \textit{et al.} \cite{liu2020GoChat} propose a  hierarchical RL method, GoChat, for topic path planning, as shown in Fig~\ref{gochat-pic}. The high-level strategies in GoChat determine sub-goals that guide the conversation towards the ultimate target and the low-level strategy generates the corresponding responses to achieve those sub-goals.

To plan topic paths from a global perspective, Yang \textit{et al.} \cite{yang2022topkg} introduce the global planning met\-hod integrated with a commonsense knowledge gr\-aph (KG). The key advancement is the introduction of a global RL framework that utilizes topic path planning on KG to guide the local response generation model toward persuasive targets, resulting in more coherent conversations.
To achieve persuasive goals more effectively, Lei \textit{et al.} \cite{Lei2022I-Pro} consider four factors (dialogue turn, goal completion difficulty, user satisfaction estimation, and cooperative degree) in the reward function. The targets of achieving persuasive targets quickly and maintaining the engagingness of users.

\subsubsection{Graph-based Planning}

\begin{figure*}[h]
    \centering
    \includegraphics[width=.8\textwidth]{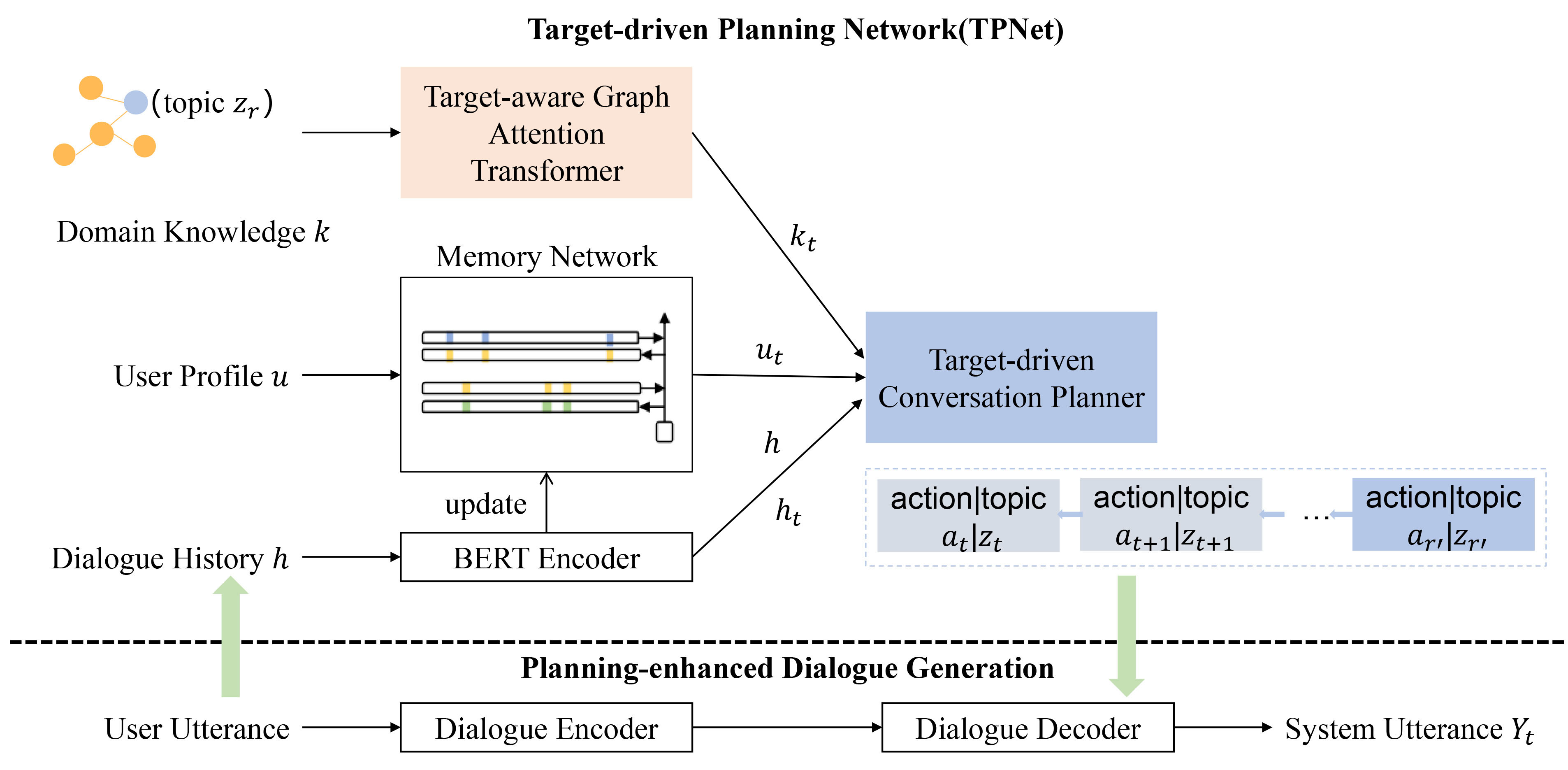}
    \caption{The overview framework of TPNet.}
    \label{TPNet-pic}
\end{figure*}

Knowledge is essential to the cognitive reasoning processes of human beings. We humans usually perform common reason during persuasive conversation to enhance the logic and persuasiveness of dialog contents. Therefore, relying on commonsense knowledge graphs for topic path planning can produce more persuasive target-related topic paths for CogAgent, thus reaching persuasive targets mo\-re efficiently.

Initially, the semantic knowledge relations amo\-ng topic keywords are captured to perform next-turn topic prediction during conversation \cite{tang2019target, qin2020dynamic}. Then the predicted topic keywords are used to retrieve appropriate candidate responses for persuasive targets.
Furthermore, Zhong \textit{et al.} \cite{zhong2021keyword} introduce commonsense knowledge graphs and Graph Neural Networks (GNN) to model the semantic relations between topic keywords and enhance the keyword-augmented response retrieval,

To plan topic paths more reasonably, Zou \textit{et al.} \cite{zou2021thinking} introduces a concept graph based on the dialogue data, where the vertices represent concepts and edges are concept transitions between utterances. The topic sequence containing multiple concepts is obtained by the multi-concept planning mo\-dule and an Insertion Transformer generates a persuasive response according to the planned topic pat\-hs.
Wang \textit{et al.} \cite{wang2023target} propose a target-driven planning network (TPNet), which models the topic path planning as a sequence generation task using Transformer, as shown in Fig~\ref{TPNet-pic}. A knowledge-target mutual attention mechanism and a set-search decoding (SSD) strategy are developed to generate topic paths based on the dialogue context.

\subsubsection{Novel Planning Mechanism}
In addition to the above research to plan topic paths in CogAgent, there are some novel planning mechanisms to be explored, summarized as follows.

Combining the strengths of multiple topic planning algorithms, the Tang \textit{et al.} \cite{tang2023eagle} propose an EAGLE model for topic path planning. Comprising a topic path sampling strategy, topic flow generator, and global planner, EAGLE achieves robustness to unseen target topics and smooth transitions. The model demonstrates enhanced global planning ability through its integrated approach, addressing limitations in existing topic-planning conversation models.

To ensure the smooth and coherent progression toward persuasive goals across different turns, Wan\-g \textit{et al.} \cite{Wang2023Brownian} introduce a consistency-driven dialogue planning approach that utilizes stochastic processes to model the temporal evolution of the conversation path dynamically. Firstly, a latent spa\-ce is defined, and Brownian bridge processes are employed to capture the continuity of goal-oriented behavior, allowing for more flexible integration of user feedback into dialogue planning, and explicitly generating conversation paths. Ultimately, these paths are employed as natural language prompts to guide the generation of persuasive dialogue.

\subsection{Argument Structure Prediction Strategy-based CogAgent}
CogAgent entails an ongoing conversation between a dialogue agent and a user at the cognitive level, where the dialogue agent proactively steers the conversation. 
As the conversation progresses, the contents presented by the dialogue agent to support its perspectives undergo dynamic transformations. Consequently, the reasonable selection and application of arguments and evidence play a pivotal role in the persuasiveness of the dialogue. 
The utilization of arguments and evidence is imperative in the process of persuasion. Firstly, employing arguments and evidence allows for the gradual decomposition and progressive reasoning of persuasive targets, thereby facilitating a logical and sequential flow of the conversation that enhances the users' acceptance of viewpoints \cite{vecchi2021towards}. Secondly, the provision of factual support elevates the credibility of persuasive discourse, thereby augmenting the persuasiveness of the conversation. 
In this section, we provide an investigation of the crucial techniques for argument mining and argument structure prediction in CogAgent, as summarized in Table~\ref{tab:Argument_CogAgent}.

\begin{table*}[htbp]
  \centering
  \caption{Representative works of argument structure prediction strategy-based cogAgent.}
  \scalebox{0.90}{
    \begin{tabular}{|m{2.8cm}<{\centering}|m{3.1cm}<{\centering}|m{12cm}<{\centering}|}
    \hline
    Solution & Work  & Description \\
    \hline
    \multirow{4}[12]{*}{\shortstack{Argument \\ mining}} & Khatib et al. \cite{al2016cross} & Classifying and structurally modeling arguments from online debate portals based on diverse vocabulary, grammar, and metric features. \\
\cline{2-3}          & Hua et al. \cite{hua2019argument} & Proposing an argument generation framework with retrieval modules and a sentence-level LSTM for generating viewpoints. \\
\cline{2-3}          & Srivastava et al. \cite{srivastava2022argument} & Using attention-based link prediction and Transformer encoder to model hierarchical causal relationships and discover associations in online argument structures. \\
\cline{2-3}          & Niculae et al. \cite{niculae2017argument} & Introducing factor graph model for argument mining, concurrently learning fundamental unit types classification and argument relationship prediction. \\
    \hline
    \multirow{4}[12]{*}{\shortstack{Argument structure \\ prediction}} & Rach et al. \cite{rach2021argument} & Proposing argument search technique using supervised learning-based relation classification to retrieve arguments for debate dialogue system \\
\cline{2-3}          & Sakai et al. \cite{sakai2020hierarchical} & Introducing an approach to consider the human agreement and disagreement, resulting in a persuasive argument with a hierarchical argumentation structure. \\
\cline{2-3}          & Prakken et al. \cite{prakken2020persuasive} & Enhancing argument modeling with a five-layer graph, serving as a knowledge base for a chatbot to identify user focal points and select rebuttal points. \\
\cline{2-3}          & Li et al. \cite{li2020exploring} & Using factor graphs to extract online debate features, incorporating them into an LSTM model to predict persuasive arguments. \\
    \hline
    \end{tabular}%
    }
  \label{tab:Argument_CogAgent}%
\end{table*}%

\subsubsection{Argument Mining} 
To integrate the argument structure into CogAgent, it is first necessary to perform argument mining according to conversation topics. Researchers embark on mining argumentative text from dialogues for CogAgent.

Debate involves the explicit use of argumentative content for dialogue expression, making it an important source of argument mining.
For example, Khatib \textit{et al.} \cite{al2016cross} utilize online debate portals to acquire both controversial and non-controversial text snippets related to several contentious topics. These snippets are organized in a semi-structured format. Eventually, by employing a diverse set of vocabulary, grammar, and metric feature types, the arguments are classified and structurally modeled.
Hua \textit{et al.} \cite{hua2019argument} propose a framework for generating arguments to opposing viewpoints. The retrieval module of this framework comprises Query Formulation, Keyphrase Extraction, and Passage Ranking and Filtering. Subsequently, a sentence-level LSTM is trained to generate a sequence of sentences.  

In online discussion platforms, people also use argumentative texts to enhance their expressions.
For instance, Tran \textit{et al.} \cite{tran2021multi} and others \cite{cheng2022iam, wang2021two, sun2021hierarchical} employ multi-task learning to unearth arguments and evidence at both the micro and macro levels, enhancing persuasive power in online discussions.
Srivastava \textit{et al.} \cite{srivastava2022argument} employs an attenti\-on-based link prediction embedding model to mod\-el the hierarchical causal relationships within common argument structures in online discussions. Th\-ey then utilize Transformer encoder layers to discover the associations and boundaries between arguments. Furthermore, they employ AMPERSAN\-D \textit{et al.} \cite{chakrabarty2019ampersand} and SMOTE \textit{et al.} \cite{chawla2002smote} to address data imbalance issues, thereby improving model accuracy. 
Furthermore, Niculae \textit{et al.} \cite{niculae2017argument} introduce a factor graph model for argument mining, wherein the mod\-el concurrently learns the classification of fundamental unit types and prediction of argument relationships. Furthermore, the parameter structures of structured SVM and RNN can enforce structural constraints (e.g., transitivity), while also representing dependencies between adjacent relationships and propositions.

\begin{figure*}[htbp]
    \centering
    \includegraphics[width=.75\textwidth]{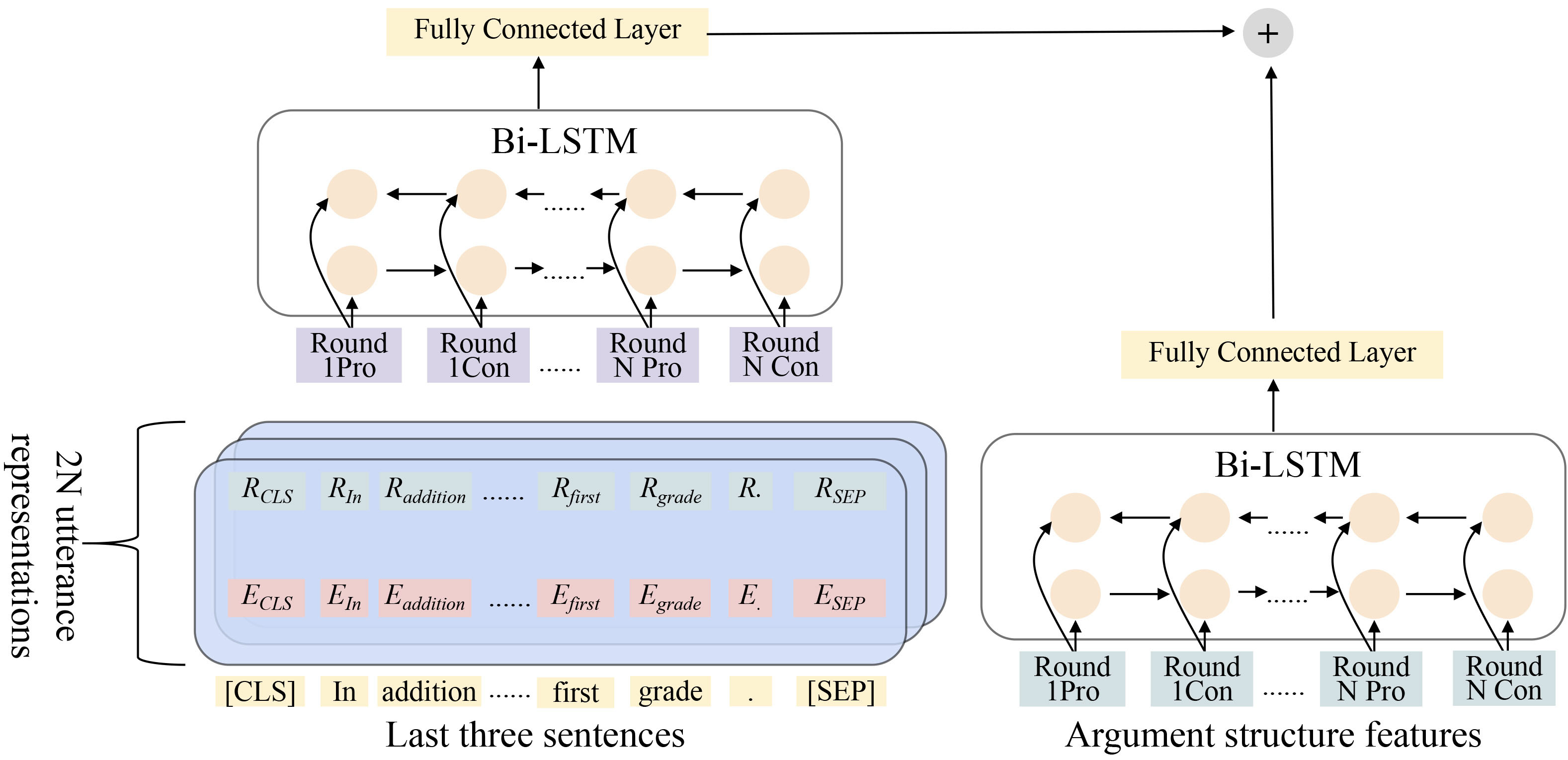}
    \caption{The overall framework of the model for predicting which side makes more convincing arguments \cite{li2020exploring}.}
    \label{li-exploring}
\end{figure*}

\subsubsection{Argument Structure Prediction} 
Dialogue systems of persuasive tasks co\-mmonly rely on structured knowledge concerning arguments and their relationships. Numerous resear\-chers have demonstrated that predicting argument structures and integrating them into CogAgent can significantly enhance topic consistency, content coherence, and persuasiveness of persuasive dialogue co\-ntents \cite{slonim2021autonomous, rach2020increasing, sakai2020hierarchical}. 

For example, Rach \textit{et al.} \cite{rach2021argument} propose an argument search technique for a debate dialogue system, which utilizes supervised learning-based relation classification to retrieve arguments mapped to a generic tree structure for the dialogue model. 
Sakai \textit{et al.} \cite{sakai2020hierarchical} introduce an approach to consider human agreement and disagreement, resulting in a persuasive argument with a hierarchical argumentation structure. The dialogue agent selects the next action based on the user's agreement or disagreement and sends the chosen action to the response generation module to generate logically consistent and persuasive dialogue. 

For more intensive argument modeling, Prakken \textit{et al.} \cite{prakken2020persuasive} equip dialogue agents with a five-layer argument graph, consisting of 1288 nodes, with an average of three counterarguments per node. This graph serves as the knowledge base for the proposed chatbot, allowing it to dynamically identify and annotate the user's focal points on the parameters, enabling the selection of appropriate rebuttal points. 
Li \textit{et al.} \cite{li2020exploring} utilized factor graph models to extract features of argument structures from online debate platforms. These features were then incorporated into an LSTM model to predict the most persuasive arguments, as shown in Fig~\ref{li-exploring}. This study proves that the consideration of argument str\-ucture plays a vital role in producing persuasive dialogue content.

\section{Datasets and Evaluation Metrics for CogAgent}
\label{chap:dataset}
\subsection{Datasets for CogAgent}
Massive data is undeniably indispensable for training high-quality CogAgent. To foster advancement in this field, numerous large-scale and high-quality datasets have been released. In this section, we categorize existing datasets by application scenarios, including psychological counseling, debate, price negotiation, persuasion for donation, and product recommendation, summarized as Table~\ref{tab:datasets}.

\begin{table*}
\centering
\caption{A review of available datasets for CogAgent.}
\scalebox{0.82}{
\begin{tabular}{m{1.5cm}<{\centering}m{3cm}<{\centering}m{14cm}}
\hline
\textbf{Scenario} & \textbf{Dataset} & \textbf{Description} \\ 
\hline
\multicolumn{1}{c}{\multirow{4}{2cm}{\textbf{Psychological counseling}}} & \textbf{ESConv \cite{liu2021towards}} & The first dataset for psychological counseling, annotated with persuasive strategies. \\ 
\cmidrule{2-3}
 & \textbf{AUGESC \cite{zheng2023augesc}} & The enhanced dataset from ESConv using LLMs with a broader range of topics. \\ 
\cmidrule{2-3}
 & \textbf{PsyQA \cite{sun2021psyqa}} & A Chinese mental health support dataset featuring annotated persuasive strategies. \\ 
\hline
\multicolumn{1}{c}{\multirow{6}{2cm}{\textbf{Debate}}} & \textbf{IAC \cite{walker2012corpus}} & Argumentative dialog dataset with curated threads, posts, and annotations. \\ 
\cmidrule{2-3}
 & \textbf{Winning Arguments \cite{tan2016winning}} & A metadata-rich subset of r/ChangeMyView subreddit conversations includes data on the success of user utterances in persuading the poster. \\ 
\cmidrule{2-3}
 & \textbf{DebateSum \cite{roush2020debatesum}} & A dataset for the competitive formal debate with corresponding argument and extractive summaries. \\ 
\hline
\multicolumn{1}{c}{\multirow{4}{2cm}{\textbf{Price negotiation}}} & \textbf{Craigslist-Bargain \cite{he2018decoupling}} & A human-human dialogue dataset for price negotiation where the buyer and seller are encouraged to reach an agreement to get a better deal. \\ 
\cmidrule{2-3}
 & \textbf{Negotiation-Coach \cite{zhou2019dynamic}} & An additional negotiation coach based on CraigslistBargain, which monitors the exchange between two annotators and provides real-time negotiation strategy. \\ 
\hline
\multicolumn{1}{c}{\multirow{7}{2cm}{\textbf{Persuasion for donation}}} & \textbf{Persuasion For Good \cite{wang2019persuasion}} & A collection of online conversations where one participant (the persuader) tries to convince the other (the persuadee) to donate to a charity. \\ 
\cmidrule{2-3}
 & \textbf{EPP4G and ETP4G \cite{mishra2022pepds}} & Datasets extending Persuasion For Good by annotating it with the emotion and politeness-strategy labels. \\ 
\cmidrule{2-3}
 & \textbf{FaceAct \cite{dutt2020keeping}} & A dataset extending Persuasion For Good by adding the utterance-level annotations that change the positive and/or the negative face of the participants in a conversation \\ 
\hline
\multicolumn{1}{c}{\multirow{7}{2cm}{\textbf{Product recommendation}}} & \textbf{TG-ReDial \cite{zhou2020towards}}& A dataset consisting of dialogues between a seeker and a recommender. \\ 
\cmidrule{2-3}
 & \textbf{DuRecDial \cite{liu2020towards}} & A human-to-human Chinese dialog dataset, which contains multiple sequential dialogues for every pair of a recommendation seeker and a recommender. \\ 
\cmidrule{2-3}
 & \textbf{INSPIRED \cite{hayati2020inspired}} & A movie recommendation dataset, consisting of human-human dialogues with an annotation scheme for persuasive strategies. \\
\bottomrule
\end{tabular}
}
\label{tab:datasets}
\end{table*}

\subsubsection{Datasets for Psychological Counseling}
Psychological counseling is a typical field of persuasive dialogue, where CogAgent reduces users' psychological anxiety and encourages positive emotions through the persuasive dialogue process. Researchers have released several datasets for psychological counseling.

\textbf{ESConv.} ESConv\footnote{\url{https://github.com/thu-coai/Emotional-Support-Conversation}} \cite{liu2021towards} is a well-designed and rich, effective corpora for psychological counseling, consisting of 1,053 dialogue pairs and a total of 31,410 sentences. Each dialogue pair includes information about the initial emotional state of the seeker, the persuasive strategies employed by the supporter during each interaction, and the content of the conversation. The dataset encompasses seven distinct emotional states and eight supportive strategies, with the labeling of these strategies being inspired by Hill's Helping Skills Theory \cite{hill2009helping}.

\textbf{AUGESC.} The limitations imposed by crowdsourcing platforms on data themes and collection methods, along with the substantial regulatory costs, have hindered the extension of downstream dialogue models to open-domain topics. In response, Zheng et al. augment ESConv to AUGESC\footnote{\url{https://github.com/thu-coai/AugESC}} \cite{zheng2023augesc} using LLMs, which comprises 65,000 dialogue sessions and a total of 1,738,000 utterances. It substantially expands the scale of ESConv and encompasses a broader range of topics.

\textbf{PsyQA.} PsyQA\footnote{\url{https://github.com/thu-coai/PsyQA}} \cite{sun2021psyqa} is a Chinese mental heal\-th support dataset collected from a Chinese mental health service platform, including 22,000 questions and 56,000 lengthy, well-structured answers. In line with psychological counseling theory, PsyQA annotates some of the answer texts with persuasive strategies and further conducts in-depth analyses of the lexical features and strategic patterns within counseling responses.

\subsubsection{Datasets for Debate}
Debates are typically persuasive scenarios in which each party of the debate organizes arguments to persuade the other party to accept his or her side's viewpoints. Existing datasets for debate are listed as follows. 

\textbf{Internet Argument Corpus (IAC).} IAC\footnote{\url{https://nlds.soe.ucsc.edu/iac}} \cite{walker2012corpus} is a scriptless argumentative dialog dataset, comprising 390,704 posts extracted from 11,800 discussions on the online debate platform \url{4forums.com}. Within this corpus, a manually curated subset of 2,866 threads and 130,206 posts is formed, categorized based on discussion topics.
Extended from IAC, IAC 2\footnote{\url{https://nlds.soe.ucsc.edu/iac2}} \cite{abbott2016internet} is a corpus for research in political debate on Internet forums, consists of three data sets: 4forums (414K posts), ConvinceMe (65K posts), and a sample from CreateDebate (3K posts).

\textbf{Winning Arguments.} To delve deeper into the mechanisms of changing others' viewpoints in social interactions, Tan et al. \cite{tan2016winning} introduce the Wining Arguments (ChangeMyView) Corpus. Wining ArgumentsCorpus is a metadata-rich subset of conversations made in the \textit{r/ChangeMyview} subreddit between 1 Jan 2013 - 7 May 2015, with information on the delta (success) of a user's utterance in convincing the poster. There are 34911 Speakers, 293297 Utterances, and 3051 Conversations.

\textbf{DebateSum.} DebateSum\footnote{\url{https://debate.cards/}} \cite{roush2020debatesum} is a dataset for the competitive formal debate, including 187,386 unique pieces of evidence with corresponding argument and extractive summaries. The argument data is collected from the National Speech and Debate Association over 7 years.

\subsubsection{Datasets for Price Negotiation}
Price negotiation is an everyday persuasive scenario where buyers and sellers reach their desired price through the persuasive dialog process. Datasets for price negotiation are summarized as follows.

\textbf{CraigslistBargain.} CraigslistBargain\footnote{\url{https://worksheets.codalab.org/worksheets/0x453913e76b65495d8b9730d41c7e0a0c/}} \cite{he2018decoupling} is a human-human dialogue dataset for price negotiation, which consists of 6682 dialogues, collected using Amazon Mechanical Turk (AMT) in a negotiation setting where two workers were assigned the roles of buyer and seller, respectively. The buyer is additionally given a target price and both parties are encouraged to reach an agreement while each of the workers tries to get a better deal.

\textbf{Negotiation-Coach.} Negotiation-Coach\footnote{\url{https://github.com/zhouyiheng11/Negotiation-Coach}} \cite{zhou2019dynamic} introduce an additional negotiation coach based on CraigslistBargain, which monitors the exchange between two annotators and provides real-time negotiation strategy recommendations to the seller for achieving better deals.

\subsubsection{Datasets for Persuasion for Donation}
Persuasion for donation is very common in life, where the persuader persuades others to donate pro\-perty or labor to charities for a public good purpose. Datasets for persuasion for donation are listed as follows.

\textbf{Persuasion for Social Good.} Persuasion for Social 
 Good\footnote{\url{https://gitlab.com/ucdavisnlp/persuasionforgood}} \cite{wang2019persuasion} is a collection of online conversations generated by AMT workers, where one participant (the persuader) tries to convince the other (the persuade\-e) to donate to a charity. This dataset contains 1017 conversations, along with demograp\-hic data and responses to psychological surveys fro\-m users. 300 conversations also have per-sentence human annotations of dialogue acts that pertain to the persuasion setting, and sentiment.

\textbf{EPP4G and ETP4G.} EPP4G and ETP4G \footnote{\url{https://github.com/Mishrakshitij/PEPDS}} \cite{mishra2022pepds} extend Persuasion For Good by annotating it with the emotion and politeness-strategy labels.

\textbf{FaceAct.} FaceAct\footnote{\url{https://github.com/ShoRit/face-acts}} \cite{dutt2020keeping} further extend Persuasion For Good by adding the utterance-level annotations that change the positive and/or the negative face of the participants in a conversation. A face act can either raise or attack the positive face or negative face of either the speaker or the listener in the conversation.

\subsubsection{Datasets for Product Recommendation}
Product recommendation intends to induce the recommended person to accept or buy a particular pro\-duct through persuasive dialogues. Datasets fo\-r pro\-duct recommendation are listed as follows.

\textbf{TG-ReDial.} TG-ReDial\footnote{\url{https://github.com/RUCAIBox/TG-ReDial}} \cite{zhou2020towards} consists of 10,0\-00 two-party dialogues between a seeker and a recommender in the movie domain. 

\textbf{DuRecDial.} DuRecDial\footnote{\url{https://github.com/PaddlePaddle/Research/tree/master/NLP/ACL2020-DuRecDial}} \cite{liu2020towards} is a human-to-human Chinese dialog dataset (about 10k dialogs, 156k utterances), which contains multiple sequential dialogues for every pair of a recommendation seeker (user) and a recommender (bot). In each dialogue, the recommender proactively leads a multi-type dialogue to approach recommendation targets and then makes multiple recommendations with ri\-ch interaction behavior.

\textbf{INSPIRED.} INSPIRED\footnote{\url{https://github.com/sweetpeach/Inspired}} \cite{hayati2020inspired} is a movie recommendation dataset, consisting of 1,001 human-human dialogues with an annotation
scheme for persuasive strategies based on social science theories.

\subsection{Evaluation metrics Toward CogAgent}

The reasonable evaluation of the quality of CogAgent is a challenging dilemma. Different from the open-domain dialog system, the evaluation of CogAgent needs to be performed under different persuasion scenarios and multifaceted persuasive goal\-s. This requires judging the quality of dialogue response while emphasizing the persuasive effects in specific persuasive contexts and assessing the adaptability and persuasiveness of the system's cognitive strategies in different domains.
Up to now, there is no unified theory on how to effectively evaluate CogAgent, and researchers predominantly employ two kinds of evaluation methods: automatic evaluation metrics and human evaluation.
We summarized commonly used automatic evaluation and human evaluation metrics in table~\ref{tab:evaluation}.
Notably, evaluating CogAgent based on LLMs has also recently received significant attention.

\begin{table*}
\centering
\caption{Evaluation metrics for CogAgent.}
\scalebox{0.8}{%
\begin{tabular}{m{2cm}<{\centering}m{1.5cm}<{\centering}m{9.5cm}<{\centering}m{7cm}<{\centering}}
\toprule
\textbf{Evaluation Method} & \textbf{Category} & \textbf{Description} & \textbf{Metrics} \\ 
\midrule
\multirow{6}{*}{\shortstack{\textbf{Automatic} \\ \textbf{evaluation}}} & Overlap-based & Measuring the degree of text
  overlap between generated responses and golden responses & BLEU \cite{papineni2002bleu}, ROUGE \cite{lin2003automatic}, METEOR \cite{lavie2007meteor}, CIDEr \cite{vedantam2015cider} \\ 
\cmidrule{2-4}
 & Embedding-based & Evaluating the semantic
  similarity of embedding vectors between generated responses and reference
  ones & \shortstack{Greedy Matching \cite{rus2012optimal}, Embedding averaging \cite{wieting2015towards}, \\ Vector Extreme \cite{forgues2014bootstrapping}} \\ 
\cmidrule{2-4}
 & Learning-based & Employing machine learning
  models to predict the quality scores of generated responses, relying not only
  on given references & ADEM \cite{lowe2017towards} \\ 
\midrule
\multicolumn{2}{c}{\textbf{Human evaluation}} & Scoring
  by human annotators to evaluate the quality of the generated responses with
  subjective judgment & Fluency, Coherence,
  Contextualization, Emotional expression, Diversity, Persuasiveness \\
\bottomrule
\end{tabular}
\label{tab:evaluation}
}
\end{table*}

\subsubsection{Automatic Evaluation Metrics}

Automatic evaluation metrics evaluate the performance of CogAgent by calculating the similarity between the responses generated by CogAgent and ground truths. There are typical categories of automatic evaluation metrics: overlap-based methods, embedding-based methods, and learning-based tec\-hniques.

\textbf{Overlap-based metrics.} 
Overlap-based methods measure the degree of text overlap between generated responses and golden responses, with pa\-rticular emphasis on the number of the same n-grams. These methods quantify the similarity of the text, especially the local structural similarity, to measure the quality of generated responses.
Classical Overlap-based methods include \textbf{BLEU} \cite{papineni2002bleu}, \textbf{ROU\-GE} \cite{lin2003automatic}, \textbf{METEOR} \cite{lavie2007meteor} and \textbf{CIDEr} \cite{vedantam2015cider}.
Amo\-ng these, BLEU evaluates response quality by comparing the harmonic mean of n-gram overlaps between generated responses and the golden ones. 
BL\-EU is a straightforward and intuitive metric, yet it is constrained by surface features and may exhibit a weak capture of semantic relevance.
ROUG\-E calculates the length of the longest common subsequences between generated and golden responses and considers the precision and recall to evaluate the quality. 
METEOR integrates multiple aspects of information, including precision, recall, and syntactic structure, providing a more comprehensive evaluation.
CIDEr evaluates the semantic similarity between generated responses and ground truths using n-gram level cosine similarity.
These metrics have been widely applied in the evaluation of open-domain dialog systems, but they focus mainly on surface features of the response and may not capture semantic relevance. In addition, relying solely on n-gram overlap to measure similarity may not always accurately evaluate the quality of long texts.

\textbf{Embedding-based metrics.} 
Embedding-based metrics evaluate the semantic similarity of embedding vectors between generated responses and reference ones.
These methods utilize pre-trained wo\-rd embedding models (e.g., BERT \cite{devlin2019bert}) to map textual responses into embedding vectors, thus capturing the semantic relationships between the texts more accurately.
Specifically, \textbf{Greedy Matching} \cite{rus2012optimal} computes the cosine similarity of word embeddings between each word in generated response and golden ones. \textbf{Embedding averaging} \cite{wieting2015towards} averages all words in the sentence to calculate the sentence-level similarity. \textbf{Vector Extrema} \cite{forgues2014bootstrapping} takes the most extreme value in the embedding vector to represent the response to be evaluated.
In essence, embedding-based metrics emphasize the semantic quality of CogAgent more than overlap-based metrics and better capture the semantic correlations between generated responses and references.

\textbf{Learning-based metrics.} Learning-based metrics employ machine learning models to predict the quality scores of generated responses, relying not only on given references but aiming to better correlate with human judgment.
ADEM \cite{lowe2017towards} is a deep model-based evaluation metric for dialogue systems.
A hierarchical RNN model is trained in a semi-supervised manner to capture semantic information and contextual associations and align with the human preferences for dialogue responses.

In summary, automatic evaluation metrics offer advantages in terms of efficiency and consistency. However, they face challenges in terms of semantic understanding, manual annotation costs, and model complexity. When selecting and applying automat\-ed evaluation metrics, it is important to balance their advantages and disadvantages according to sp\-ecific persuasive tasks and scenarios.

\subsubsection{Human Evaluation} 
Human evaluation involves subjective judgment an\-d scoring by human annotators to evaluate the quality of the generated responses.
The annotators are usually domain experts and crowd workers who subjectively evaluate the generated responses based on specified criteria and task requirements.
Compared to automatic evaluation metrics, human evaluation captures the subjectivity, emotion, and use of persuasive strategies expressed by CogAgent. 
Therefore, the flexibility and highly customizable nature of human evaluation becomes a reliable means to ensure that the quality of CogAgent is robustly evaluated.

The human evaluation mainly evaluates CogAgent in the following main aspects: fluency, coherence, contextualization, emotional expression, diversity, and persuasiveness. 
In summary, human evaluation has advantages in terms of insightful and accurate evaluation of the quality of CogAgent. Ho\-wever, it also has limitations in terms of cost and efficiency, due to the requirement of human labor and time resources. Therefore, in practical applications, researchers need to strike a balance between human and automatic evaluation and choose the evaluation metrics that best suit the task requirements.

\section{Open Issues and Future Trends}
\label{chap:open}
Though researchers have made considerable efforts to address the above challenges in CogAgent, there are still open issues to be resolved. In this section, we present some open issues and future development trends for CogAgent to promote the advancement of the research community.

\subsection{Comprehensive Modeling of Cognitive Psychology Theory for CogAgent}
Although we have summarized some of the cognitive psychology theories, a comprehensive investigation of the cognitive mechanisms of persuasive dialogues from a cognitive psychology perspective is essential for understanding users' cognitive weaknesses and generating engaging persuasive dialogu\-es.
Many researchers have demonstrated the indispensability of employing specific strategies to achieve persuasive effects based on different cognitive psychology theories. Utilizing cognitive strategies, CogAgent can avoid cognitive dissonance in users and efficiently persuade them to accept specific viewpoints \cite{wang2019persuasion, chen2019multi, yang2021improving}.
Prakken \textit{et al.} \cite{jing2020activating} argue that psychological dissonance occurs when individuals are confronted with multiple conflicting cognitions. To alleviate this dissonance, three approaches can be used: changing cognitively relevant factors in the environment, introducing new cognitive elements, and changing cognitive elemen\-ts in behavior. CogAgent should be aware of cognitive dissonance to mitigate the obstacles it creates in the persuasion process.
In addition, researchers utilize the dual process theory of persuasion and guide the persuasive process with the Elaboration Likelihood Model (ELM) \cite{petty2015emotion}, a theory that focuses on cognitive and affective appeals in persuasion.
Another noteworthy aspect is modeling the user's cognition. Proposing agreements or making concessions promptly facilitates the perception of the user's cognitive state, enabling CogAgent to adapt to changes in the user's cognition on time and avoiding the failure of the persuasive process \cite{thimm2014strategic}.

Besides using data analysis to study the mechanisms of persuasive dialog, we can also explore this phenomenon from the perspective of the cognitive functions of the human brain. Advances in neuroscience have provided valuable methods for studying the cognitive mechanisms of persuasive dialogue.
As Poldrack \textit{et al.} state \cite{poldrack2015progress}, the use of electroencephalography (EEG), magnetoencephal\-ography (MEG), functional magnetic resonance im\-aging (fMRI), and other brain-imaging tools can deepen our understanding of how the human brain produces social behavior. Arapakis \textit{et al.} \cite{arapakis2017interest} use brainwave recordings to measure users' interest in news articles, and the experimental results suggest that frontal asymmetry (FFA) can objectively assess users' receptive preferences for content.
Exploring the changes in neural signals in the brain of the persuadee during persuasive conversations to model which persuasive factors are effective in being accepted by users and convincing them to adopt persuasive targets is a promising research direction.

\subsection{Model Adaptivity/Generality of CogAgent}
Equipping CogAgent with cross-domain understan\-ding and generation capabilities is a promising research direction. Existing CogAgent usually focuses on one specific persuasion scenario, such as persuasion for social good, bargaining, and debating. 
However, it is crucial to develop the ability of CogAgent to understand and transfer through multiple domains, which enables CogAgent to dynamically optimize cognitive strategies based on different persuasive targets and efficiently perform persuasive tasks.
For example, Wolf \textit{et al.} \cite{wolf2019transfertransfo} utilize transfer learning to jointly fine-tune multiple unsupervised response prediction tasks. They demonstrate the effectiveness of language model transfer learning on the PERSONA-CHAT dataset, especially on the dialogue response generation task.
Qian \textit{et al.} \cite{qian2019domain} propose a meta-learning-based approach to domain adaptive dialogue generation that learns from multiple resource-rich tasks. They utilize multiple resource-rich single-domain dialog dat\-asets to train the dialogue system so that it can adapt to new domains with minimal training samples.
Therefore, improving the transferability of CogAgent across different domains using transfer learning and other advanced approaches is an important step towards the universal CogAgent.

\subsection{Multi-party CogAgent}
Existing research of CogAgent has demonstrat\-ed remarkable performance in two-party conversat\-ion\-al scenarios.
However, in real world, multi-party conversations (MPCs) are more prevalent and require CogAgent to persuade multiple participants simultaneously.
Unlike existing persuasive dialog systems, multi-party dialogue scenarios require the collaboration of multiple CogAgent to efficiently achieve persuasion targets \cite{shi2019deep, ju2022learning,yuan2024communication}. 
Specifically, a single CogAgent is prone to be overly purposeful when interacting with users, which can cause the users' resentment and resistance and hinder the realization of persuasion targets. In contrast, multiple CogAgents can assume different persuasive roles, cooperate, and persuade from different perspectives, thus winning users' trust and realizing persuasion targets more effectively.
Existing studies have explored MPCs in open-domain dialogue systems. 
For instance, Ito \textit{et al.} \cite{ito2022predicting} construct a multi-modal and multi-party model based on GRU to predict the persuasiveness of multiple members within a group during multi-party conversations, thereby providing a model paradigm for the study of multi-party dialogues. 
Gu \textit{et al.} \cite{gu2020speaker} propose a Speaker-Aware BERT (SABERT) model to select appropriate speaking targets from multiple users based on dialogue contexts.
Gu \textit{et al.} \cite{gu2023gift} explore the problem of "who says what to whom" in MPCs and propose a plug-and-play graphically-induced fine-tuning (GIFT) module for tuning a variety of PLMs for generalized multi-party conversation understanding.
Inspired by multi-party dialogue research, it is promising to utilize multiple CogAgents to collaborate on persuasive tasks to enhance the credibility and efficiency of the persuasion process.
Multiple CogAgents utilize persuasive roles with complementary capabilities, strategies, and trust-building to enhance persuasion and effectiveness, thereby facilitating more persuasive and successful persuasion results.

\subsection{Interpretability of Persuasive Process}
Interpretability of models can improve their credibility. Improving the interpretability of the persuasion process is essential to ensure that persuasive dialogue contents produced by CogAgent are accepted and adopted.
In recent years, the field of Natural Language Processing (NLP) has increasingly focused on improving the interpretability of deep models \cite{belinkov2020interpretability, jacovi2020towards}.
For example, Gaur \textit{et al.} \cite{gaur2021semantics} argue that domain-specific knowledge helps to understand how deep models work. They demonstrate the utility of incorporating knowledge-infuse\-d learning in knowledge graph format into complex neural networks to achieve model interpretability.
Similarly, Yasunaga \textit{et al.} \cite{yasunaga2021qa} demonstrate model interpretability and structure inference by combining a pre-trained language model a knowledge grap\-h, and a quality assurance context into a unified graph.
Currently, research on the interpretability of the persuasion process still lacks an overall framework. For the interpretability of the persuasion process, the effectiveness of the persuasion strategy can be verified from the cognitive theory, combined with the knowledge graph reasoning, and the persuasion behavior can be analyzed interactively.

\subsection{Multimodal CogAgent}
Multimodal perception and comprehension capabilities are essential for human beings in daily conversations. By understanding the multimodal surroundings around them, including visual, textual, auditory, and other modal information, we humans can produce engaging dialogues to communicate messages, emotions, and attitudes with others \cite{quek2002multimodal, turk2014multimodal}. 
Despite the outstanding natural language understanding and generation capabilities, perceiving and understanding multimodal context information is essential for natural and harmonious human-mac\-hine conversation systems \cite{jaimes2007multimodal, baltruvsaitis2018multimodal}. 
To persuade people to change their thoughts, opinions, or attitudes, it is crucial to understand the multimodal surroundings of users. Different environments may lead users to develop different attitudes towards thi\-ngs. Combining multimodal contextual information, persuasive dialogue systems can comprehensively understand users' mental states to generate more specific persuasive dialogue content.
There has been extensive research on multimodal dialogue systems that enable the understanding of image or video content through dialogue \cite{qi2020two, alamri2019audio, wu2023visual}. 
For example, Murahari \textit{et al.} adapt ViLBERT \cite{lu2019vilbert} to achieve multi-turn image-based dialogue, which understands the image information through image-text pre-trained on multimodal datasets.
Visual ChatGPT \cite{wu2023visual} integrates ChatGPT with visual foundation models to achieve visual dialogue. Different kinds of visual information, such as images, depth images, and mask matrices, are converted into language formats based on visual foundation models and the prompt manager. Then ChatGPT takes the information from visual and textual modalities to generate dialogue responses.
These efforts have laid a solid foundation for multimodal persuasive dialogue systems. The integration of multimodal information to generate more persuasive conversational content is a highly promising research direction.

\subsection{Data and Model Co-Optimization for CogAgent}
The huge impact of LLMs (e.g., ChatGPT) in the field of dialog systems has sparked the enthusiasm of researchers and has been widely used in many domains \cite{pal2023domain, liang2023code, wen2023empowering}.
For example, Liang \textit{et al.} \cite{liang2023code} rewrite the policy code for controlling a robot using LLMs. The policy code can receive and understand commands and then outputs the execution code to the API to achieve coherent control of the robot's actions through the classical chain logic.
Similarly, Wen \textit{et al.} \cite{wen2023empowering} combine the common-sense knowledge implicit in LLMs with the domai\-n-specific knowledge of mobile applications to realize hands-free speech-based interaction between users and smartphones.
LLM can be surprisingly useful in a variety of domains.
To develop a high-quality CogAgent, we can utilize LLMs to generate large-scale persuasive dialogue data to quickly validate the algorithm at an early stage. Since the capability of LLMs stems from massive amounts of data, retraining this data is hugely expensive. 
Therefore, the persuasion process also needs to be modeled to efficiently and accurately perform the persuasion task. The combination of data-driven LLMs and model-driven persuasion process is the most efficient way to develop intelligent CogAgent.
Future research directions for combining LLMs and model-driven persuasion processes include issues such as when to employ the generation abilities of LLMs, when mo\-del constraints are needed, and the rules and timing of collaboration between LLMs and the persuasion process.

\subsection{Construction of standardized datasets and be\-nchmarks}
Despite the significant progress researchers have made in CogAgent, datasets, and benchmarks for the study of CogAgent are still scarce.
The relatively small size of many existing datasets (e.g., Persuasion for good \cite{wang2019persuasion}) limits the performance of the model in a wider range of applications. The limited amount of data hinders the ability to capture the full complexity and diversity of persuasive dialogue.
Moreover, the lack of detailed annotations about cognitive strategies in existing datasets creates challenges for training persuasive dialogue agents.
Building large-scale, high-quality datasets of persuasive dialogues with rich cognitive strategy annotations is indispensable for the development of CogAgent.
Combining the superior text generation capabilities of LLMs \cite{kim2022soda, zheng2022augesc} is a potential way to build large-scale and high-quality datasets for CogAgent.

\section{Conclusion}
\label{chap:conclusion}
Persuasion is an essential ability in human social communication, and people often skillfully persua\-de others to accept their standpoints, views, or perspectives for various purposes. Consequently, persuasive dialogue systems have become an engaging research direction. In this paper, we have made a systematic survey of CogAgent. We first present some representative cognitive psychology theories to guide the design of CogAgent at the principle level and formalize the necessary cognitive strategies for generating highly persuasive dialogue contents, including the persuasion strategy, the topic path planning strategy, and the argument structure prediction strategy. Based on the formalized definition and generic architecture of CogAgent, we comprehensively investigate representative works by categorizing cognitive strategies. The available datasets and evaluation metrics for CogAgent are also summarized. Despite significant progress, the research of CogAgent is still in the early stage and massive open issues and prospective future trends to be explored, such as model adaptivity/generality of CogAgent, multi-party CogAgent, and multimo\-dal CogAgent.

\begin{acknowledgement}
This work was partially supported by the National Science Fund for Distinguished Young Scholars(62025205), and the National Natural Science Foundation of China (No. 62032020).
\end{acknowledgement}

\bibliographystyle{fcs}
\bibliography{CogAgent}

\begin{biography}{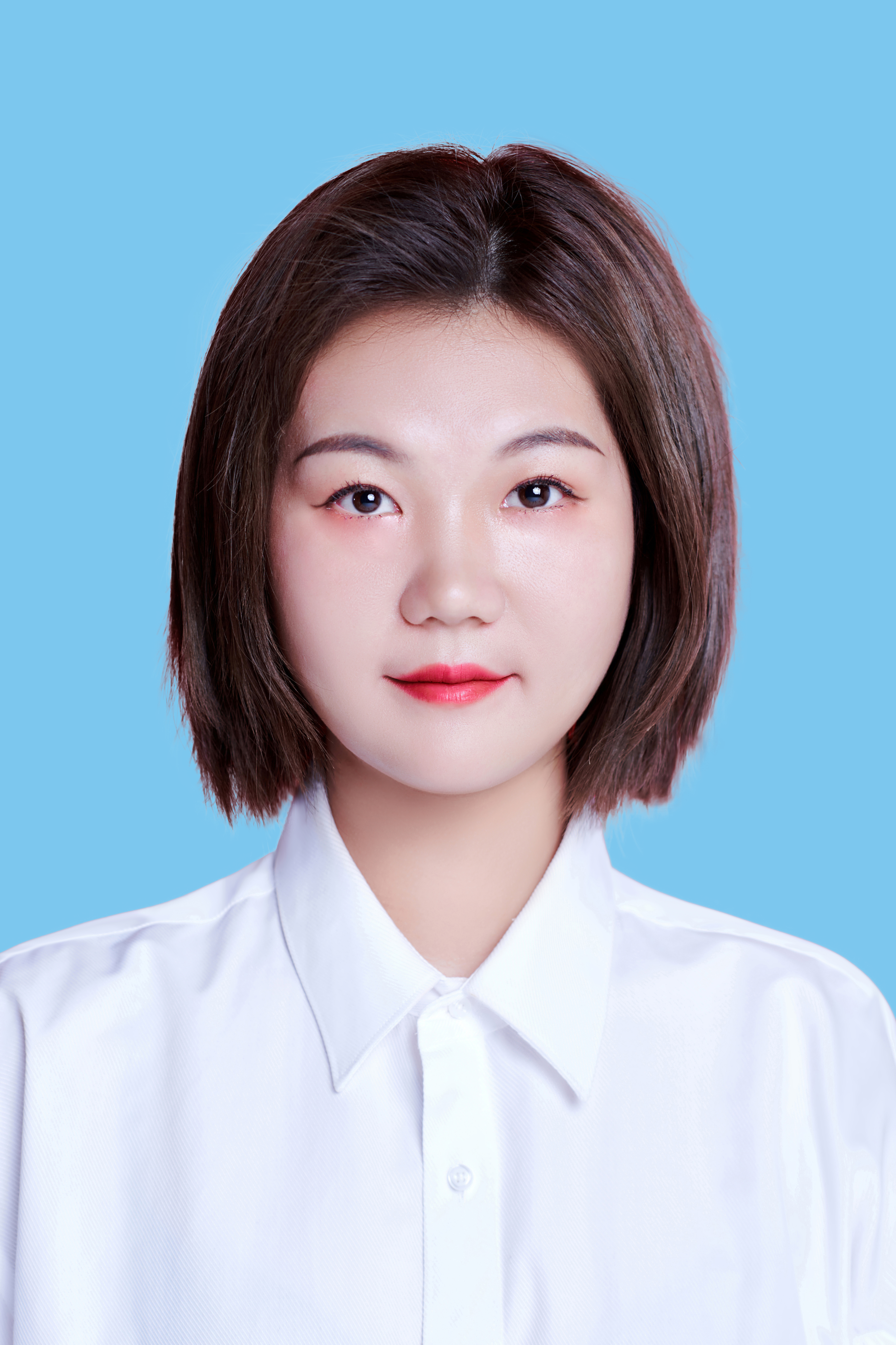}
Mengqi Chen was born in 1997. She received her master's degree in digital textiles from Xi'an Polytechnical University (XPU) in 2022. She is currently working toward a Ph.D. degree at Northwestern Polytechnical University (NWPU). Her current research interests include natural language processing, dialog systems, and large language models.
\end{biography}
\begin{biography}{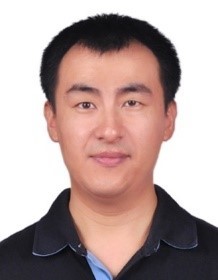}
Bin Guo was born in 1980. He is a Ph.D. professor and Ph.D. supervisor at Northwestern Polytechnical University (NWPU). He is a senior member of the China Computer Federation. His main research interests include ubiquitous computing, social and community intelligence, urban big data mining, mobile crowdsensing, and human-computer interaction.
\end{biography}
\begin{biography}{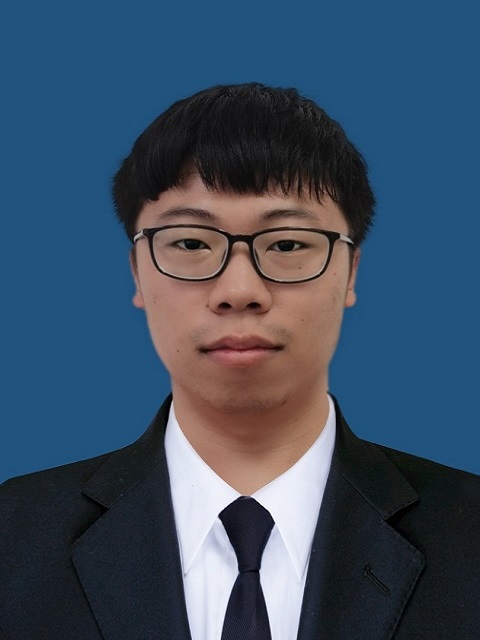}
Hao Wang was born in 1996. He received his B.E. degree in computer science and technology from Northwestern Polytechnical University (NWPU) in 2019. He is currently working toward a Ph.D. degree at NWPU. His current research interests include natural language processing, dialog systems, and large language models.
\end{biography}
\begin{biography}{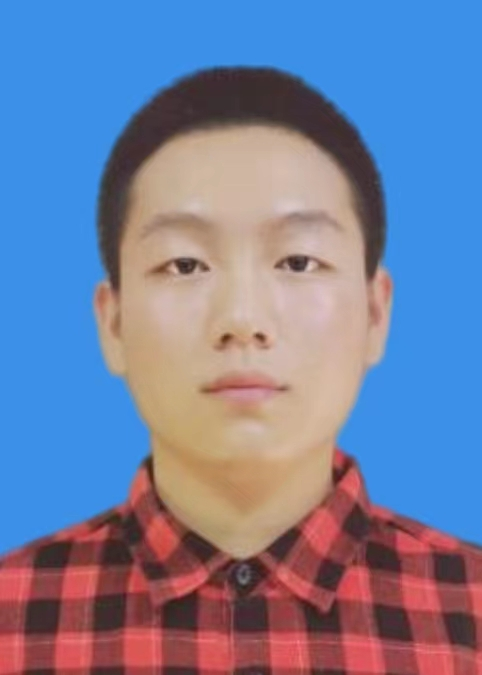}
Haoyu Li was born in 2002. He received his B.E. degree in computer science and technology from Northwestern Polytechnical University (NWPU) in 2023. He is currently working toward a master's degree at NWPU. His current research interests include natural language processing, large language models, and robot dynamic obstacle avoidance.
\end{biography}
\begin{biography}{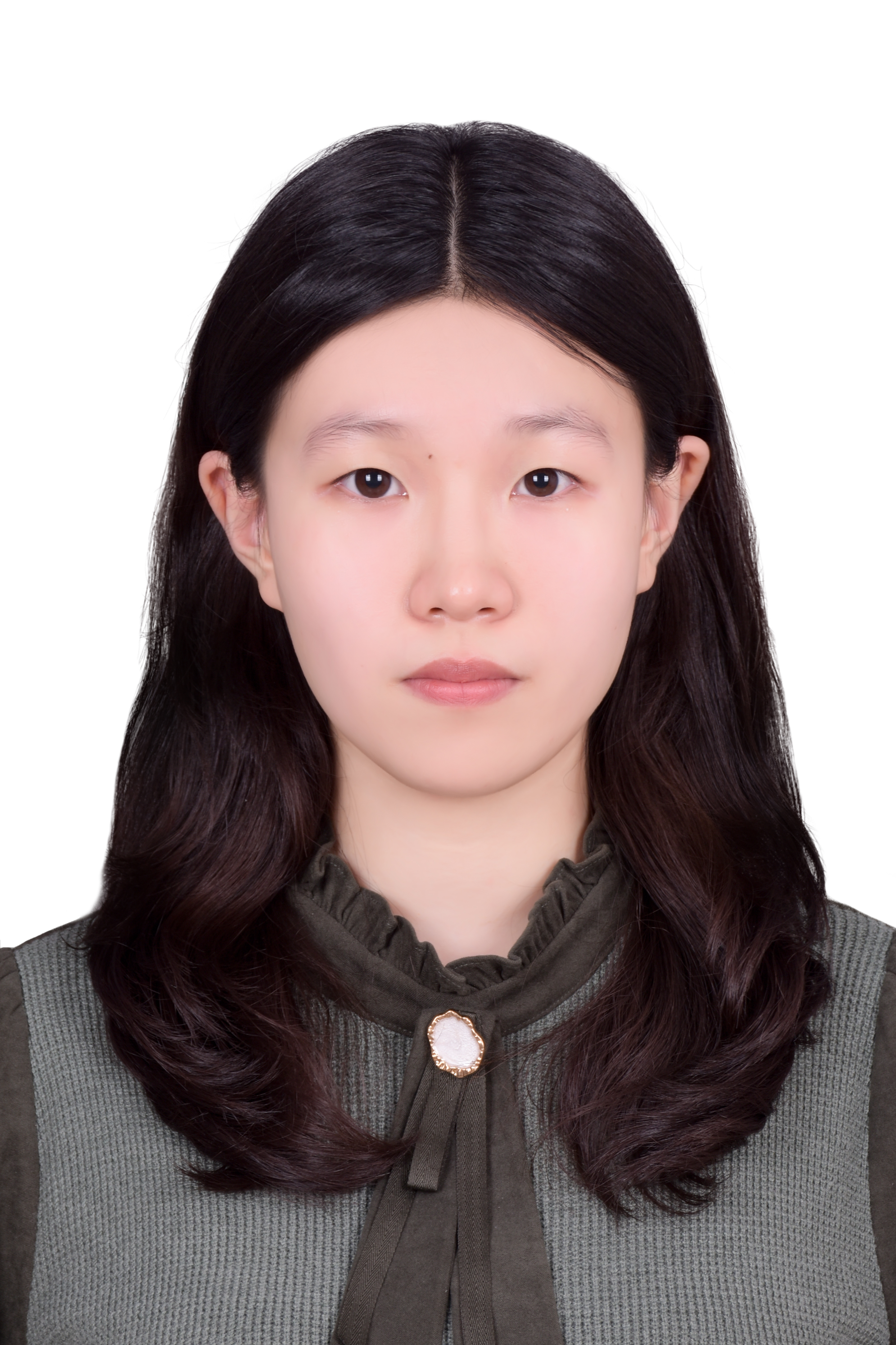}
Qian Zhao was born in 2001. She received her B.E. degree in Internet of Things engineering from Tianjin University of Technology (TUT) in 2023. She is currently working toward a master's degree at Northwestern Polytechnical University (NWPU). Her current research interests include multimodal dialogue, large language models, and visual human-computer interaction.
\end{biography}
\begin{biography}{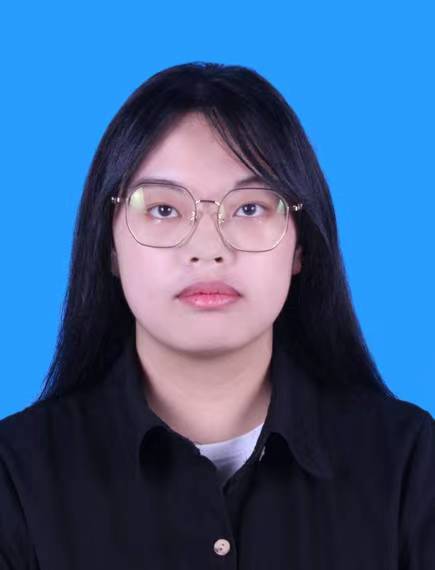}
Jingqi Liu was born in 2002. She entered Northwestern Polytechnical University(NWPU) to study for a bachelor's degree in information and computing science in 2020. Her current research interests include natural language processing, dialogue systems, and large language models.
\end{biography}
\begin{biography}{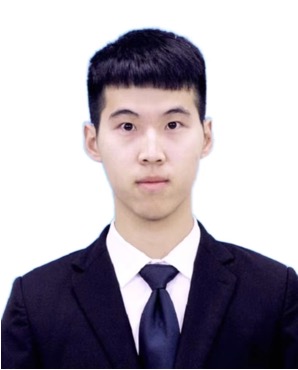}
Yasan Ding was born in 1995. He received his B.E. degree in computer science and technology from Northwestern Polytechnical University (NWPU) in 2018. He is currently working toward a Ph.D. degree at NWPU. His current research interests include fake news detection and natural language processing.
\end{biography}
\begin{biography}{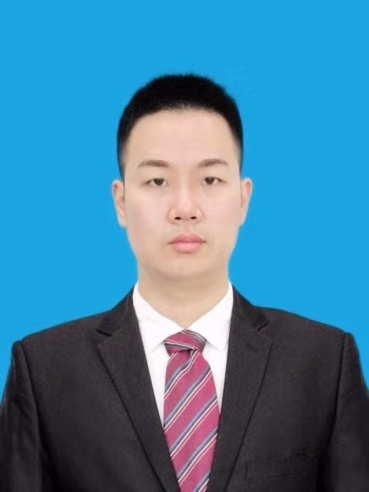}
Yan Pan was born in 1991. He is a lecturer at the Science and Technology on Information Systems Engineering Laboratory. He respectively received the B.S. degree in 2013 and the Ph.D. degree in 2020 from Northwestern Polytechnical University (NWPU). His research interests include Big Data, Machine Learning, and Crowd Intelligence.
\end{biography}
\begin{biography}{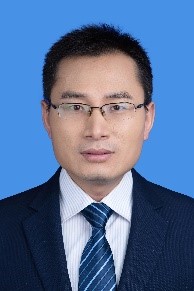}
Zhiwen Yu was born in 1977. He is a Ph.D. professor and Ph.D. supervisor. He is a senior member of the China Computer Federation. His main research interests include mobile internet, ubiquitous computing, social and community intelligence, urban big data mining, mobile crowdsensing, and human-computer interaction.
\end{biography}

\end{document}